\documentclass[conference]{IEEEtran}

\ifCLASSINFOpdf
\else
\fi
\usepackage{tikz}
\usepackage{amsmath,amsfonts,amssymb,graphicx}
\usepackage{wasysym}
\usepackage{footnote}
\usepackage{multirow}
\usepackage{pifont}
\usepackage{subfigure}
\usepackage[lined,ruled,commentsnumbered]{algorithm2e}
\usepackage{algpseudocode}
\usepackage{pdfpages}
\usepackage{tabularx}
\usepackage{booktabs}
\usepackage{tcolorbox}
\usepackage{hyperref}
\usepackage{makecell}
\usepackage{float}
\usepackage{xcolor}
\usepackage{array}
\usepackage{bbm}
\usepackage{enumitem}
\usepackage{threeparttable}
\usepackage{subcaption}
\usepackage{cite}

\begin{document}
%
\title{LLM4Trust: Exploring the Capabilities of Large Language Models for Trust Evaluation}

	

\author{
\IEEEauthorblockN{Jie Wang\IEEEauthorrefmark{1}\IEEEauthorrefmark{2},
Yanbo Sun\IEEEauthorrefmark{1},
Zheng Yan\IEEEauthorrefmark{1}\IEEEauthorrefmark{2}\textsuperscript{\ding{41}},
Jiahe Lan\IEEEauthorrefmark{1}, 
Elisa Bertino\IEEEauthorrefmark{3}}
\IEEEauthorblockA{\IEEEauthorrefmark{1}State Key Laboratory of Integrated Services Networks, School of Cyber Engineering, Xidian University}
\IEEEauthorblockA{\IEEEauthorrefmark{2}Hangzhou Institute of Technology, Xidian University, \IEEEauthorrefmark{3}Department of Computer Science, Purdue University}
\IEEEauthorblockA{\{jwang1997, yanbo.sun\}@stu.xidian.edu.cn, zyan@xidian.edu.cn, jhlan16@stu.xidian.edu.cn, bertino@purdue.edu}
}


\IEEEoverridecommandlockouts
\makeatletter\def\@IEEEpubidpullup{6.5\baselineskip}\makeatother
\IEEEpubid{\parbox{\columnwidth}{
		Network and Distributed System Security (NDSS) Symposium 2027\\
		22--26 March 2027, Seoul, Republic of Korea\\
		ISBN 978-1-970672-09-1\\  
		https://dx.doi.org/10.14722/ndss.2027.231701\\
		www.ndss-symposium.org
}
\hspace{\columnsep}\makebox[\columnwidth]{}}

\maketitle

\begin{abstract}
Trust evaluation plays a critical role in cybersecurity by supporting risk mitigation and decision-making. A variety of trust evaluation methods have been proposed, with learning-based approaches offering high accuracy and automation. However, they often require substantial ground truth, suffer from low training efficiency, lack support for basic trust properties, and provide limited explainability. Large Language Models (LLMs) offer a compelling alternative due to their strong zero-/few-shot reasoning abilities and broad knowledge. To this end, we propose LLM4Trust, the first benchmark framework that systematically explores the capabilities of LLMs for trust evaluation. We first construct diverse trust graphs to model five basic trust properties and design corresponding property understanding tasks. We then assess the ability of eight representative LLMs to understand these properties under nine prompt methods. Based on this exploration, we identify the most effective LLM-prompt combinations and apply them to five real-world datasets for validating LLMs' trust evaluation capability. During this process, we propose two strategies to extract key information from large-scale trust graphs, addressing the context window limitations of LLMs. Extensive experiments show that LLMs can effectively understand basic trust properties and have great potential for real-world trust evaluation, particularly under limited supervision. However, they remain vulnerable to attacks targeting trust graphs and demonstration examples used in few-shot prompting, and incur high inference costs. Accordingly, we propose a defense mechanism and batch inference to improve the robustness and efficiency of LLM-based trust evaluation. The source code of LLM4Trust is available at~\url{https://github.com/Jieerbobo/LLM4Trust}.
\end{abstract}


%
\IEEEpeerreviewmaketitle

\section{Introduction}
Trust is a complex and multifaceted concept, defined as the subjective perception that one entity holds towards another within a specific context. It exhibits several unique properties, including subjectivity, dynamicity, context-awareness, asymmetry, conditional transitivity, and composability~\cite{yan2008trust,sherchan2013survey,wang2022survey}. Trust evaluation is a method that quantifies trust by considering various trust-related factors. It has been widely applied to different fields for diverse purposes~\cite{wang2019graph}, e.g., fraud detection in financial and social networks~\cite{wen2024ta,fang2022integrating}, intrusion detection in communication networks~\cite{li2021surveying}, and access control in cloud computing~\cite{yan2015flexible}. In cryptocurrency markets, for instance, where illicit activities are prevalent (with an estimated \$24.2 billion transferred to illicit wallet addresses linked to criminal behavior in 2023~\cite{chainalysis2024crypto}), trust evaluation helps identify suspicious entities and reduce security risks. Overall, it provides valuable support for risk mitigation, decision-making, and system security enhancement~\cite{wang2024trustguard}.

A large number of trust evaluation methods have been proposed, which can be broadly categorized into statistics-based, inference-based, and learning-based approaches~\cite{wang2022survey}. Statistics-based methods evaluate trust by computing weighted sums of trust-related factors, while inference-based methods represent trust through multiple dimensions and infer trust using predefined rules. Although simple and intuitive, both methods heavily rely on weight selection and expert knowledge, limiting their generality across different domains~\cite{wang2024trustguard}. Learning-based methods address this limitation by using Machine Learning (ML) to automatically learn weights or rules from large amounts of trust-related data, providing automated evaluation with superior performance~\cite{lin2020guardian}. However, they still face several limitations: (i) They require substantial ground-truth labels for training, which are difficult to obtain in practice~\cite{hou2022handling}. (ii) Model training is typically computationally expensive. (iii) They struggle to capture all fundamental properties of trust~\cite{luo2025graph}, resulting in sub-optimal performance. (iv) Most learning-based methods operate as black boxes, offering no explanation for each evaluation result, thereby undermining user acceptance~\cite{han2023anomaly}. 

Recently, Large Language Models (LLMs) have achieved remarkable success across various domains, including anomaly detection~\cite{yang2024ad,zhu2024llms}, harmful content detection~\cite{thomas2025supporting,vishwamitra2024moderating}, and vulnerability management~\cite{liu2024exploring,li2025sv}. Encouraged by these advances, we ask the following question: \textbf{Can LLMs be applied to trust evaluation?} We have some intuitions supporting this possibility: (i) LLMs exhibit strong zero-shot, few-shot, and in-context learning capabilities~\cite{weiemergent,brown2020language}, which allow them to perform trust evaluation with minimal examples. This would not only reduce the reliance on large amounts of ground truth, but also avoid re-training when dealing with new tasks. (ii) LLMs possess extensive knowledge bases~\cite{wang2025the} that likely encompass trust-related concepts and properties, making them well suited to trust evaluation. (iii) LLMs demonstrate strong reasoning abilities, enabling them to understand and handle complex trust relationships between entities. (iv) LLMs support step-by-step reasoning, making it possible to explain how a trust evaluation result is derived. Therefore, LLMs have the potential to complement existing learning-based methods by alleviating some of their limitations.

To systematically investigate the above question, we decompose it into two specific research questions: \textbf{RQ1}: Do LLMs have an inherent understanding of trust properties? \textbf{RQ2}: Can LLMs perform trust evaluation, and how does their performance compare with state-of-the-art methods?
By addressing \textbf{RQ1}, we examine whether LLMs can effectively understand the basic properties of trust using different prompts, thereby revealing their underlying knowledge of trust-related concepts. Addressing \textbf{RQ2}, on the other hand, allows us to assess the practical effectiveness of LLMs in performing trust evaluation and to benchmark their performance against existing methods. 

However, answering these two questions is non-trivial due to the following challenges:
\textbf{C1}: Trust is a complex, multifaceted concept with diverse properties, yet existing datasets lack explicit annotations or structured representations of these properties. This absence makes it difficult to construct evaluation tasks that can comprehensively assess LLMs' understanding of basic trust properties.
\textbf{C2}: LLMs are constrained by context window sizes (i.e., the maximum number of tokens they can process during inference), making it challenging to include all relevant information in a single prompt. This limitation may result in incomplete reasoning or degraded performance in complex trust evaluation scenarios.

To address these challenges, we propose LLM4Trust, the first benchmark framework that systematically explores the capabilities of LLMs for trust evaluation.
\textbf{To address C1}, we focus on graph-structured data, as trust relationships between entities in different networks can be naturally modeled as graphs, with nodes representing entities and edges representing their trust relationships. This graph-based abstraction enables us to easily model diverse trust properties and design evaluation tasks accordingly. Specifically, we construct trust graphs~\footnote{A trust graph is a graph-based representation of trust relationships between a set of entities.} to explicitly model distinct properties of trust using three data generators: the Erdős-Rényi model~\cite{erd6s1960evolution}, the Stochastic Block model~\cite{holland1983stochastic}, and the Forest Fire model~\cite{leskovec2007graph}. These generators are selected to ensure structural diversity and representativeness of the constructed trust graphs. Based on these graphs, we design a number of question-answer pairs regarding five trust properties (i.e., dynamicity, asymmetry, conditional transitivity, composability, and context-awareness), referred to as property understanding tasks. To further enhance task diversity and difficulty, we vary graph parameters, such as time span, graph size, graph density, and trust level range. We then evaluate eight popular LLMs (i.e., GPT-3.5, GPT-4o, DeepSeek-V3~\cite{liu2024deepseek}, Qwen-2.5-Max~\cite{qwen25}, Llama-4-Scout, Llama-4-Maverick, Claude-3.7-Sonnet, and GPT-5) on these tasks using nine prompt methods (i.e., 0-shot, 1-shot, few-shot, knowledge, role, chain-of-thought, and three combinations of these), aiming to identify the most effective LLM-prompt combinations for trust property understanding.

\textbf{To address C2}, we propose two strategies to extract key information for a given trustor-trustee pair (i.e., two nodes with a trust relationship). First, we construct a subgraph that captures the most relevant trust information by including all nodes and edges within a specified hop range around the node pair, subject to a size constraint. Second, for dynamic trust graphs where trust relationships are timestamped, we prioritize recent relationships based on the assumption that recent behaviors better reflect current trust states than outdated ones~\cite{lin2021medley,wang2025cat}. To determine the optimal hop range and subgraph size (i.e., the best input format), we conduct a parameter study on small-scale real-world datasets for saving LLM querying costs. Using the best input format, we evaluate the practical effectiveness of the best-performing LLM-prompt combinations (identified in \textbf{RQ1}) for trust evaluation on five large-scale real-world datasets~\cite{massa2009bowling,rossi2015network,kumar2016edge,kumar2018rev2,tang2015trust}. Their performance is further compared with ten representative trust evaluation methods~\cite{wang2024trustguard,lin2020guardian,lin2021medley,liu2017opinionwalk,yao2013matri,liu2019neuralwalk,jiang2022gatrust,huo2024trustgnn,massa2005controversial,wen2023dtrust}. In addition, we assess the robustness of LLMs against attacks targeting both trust graphs and demonstration examples used in few-shot prompting.

Our evaluation and analysis yield the following key findings:
(i) LLMs can effectively understand basic trust properties. In particular, the best-performing LLM-prompt combinations achieve over 97\% accuracy across all property understanding tasks, significantly outperforming baselines based on random guessing or heuristics.
(ii) The difficulty of understanding trust properties increases as graph size and time span expand, while remaining relatively unaffected by changes in data generators, trust level range, and graph density. This highlights the challenges of applying LLMs to real-world trust graphs that are both large-scale and dynamic.
(iii) LLMs outperform non-learning-based approaches but remain less accurate than the state-of-the-art learning-based methods when sufficient labels are available. Under limited supervision, however, LLMs show clear advantages over existing methods, demonstrating their strong trust evaluation capabilities with minimal task-specific supervision.
(iv) LLMs offer a transparent, step-by-step inference process that enhances the explainability of trust evaluation results and fosters user acceptance.
(v)~Despite their promise, LLMs remain vulnerable to attacks on trust graphs, are highly sensitive to the quality of demonstration examples, and incur high inference costs. 

\textbf{To address the new challenges identified in finding~(v)}, we further propose a defense mechanism that integrates adversarial example augmentation, temporal edge filtering, and consistency correction. This defense effectively mitigates the impact of the above two attacks, recovering performance by up to 29.08\%. To improve efficiency, we adopt batch processing for parallel inference, which significantly reduces inference time per evaluation instance at the cost of some degradation in accuracy and explainability.

To summarize, we make the following contributions:
\begin{itemize}[leftmargin=*]
    \item We propose LLM4Trust, the first benchmark framework to systematically assess LLMs' capabilities for trust evaluation. 
    \item We design five property understanding tasks, considering three data generators and four graph parameters, to comprehensively evaluate eight LLMs' understanding of trust properties under nine prompt methods.
    \item We propose two strategies to address the context window limitations of LLMs and compare their trust evaluation performance against ten representative methods on five real-world datasets.
    \item Based on the experimental results, we identify key challenges regarding robustness and scalability in LLM-based trust evaluation and propose corresponding solutions.
\end{itemize}

\section{Background and Related Work}
In this section, we first introduce the basics of trust and trust evaluation. We then briefly review existing trust evaluation methods and recent efforts to assess LLMs' knowledge and capabilities across different domains.


\subsection{Trust and Trust Evaluation}
\textbf{Trust.} \label{trust_basics}
Trust is a complex concept characterized by several distinct properties~\cite{sherchan2013survey,luo2025graph}:
(i) \textit{Subjectivity:} Trust is inherently subjective; that is, different nodes may have different levels of trust towards the same target.
(ii) \textit{Dynamicity:} Trust changes with new interactions and usually decays over time. Recent interactions typically carry more weight than older ones when evaluating trust~\cite{lin2021medley}.
(iii) \textit{Context-awareness:} Trust varies across different contexts. A context refers to any information that describes a specific situation in which trust is established. For example, node $u$ may trust node $v$ in context $c_1$ (e.g., coding ability) but not in context $c_2$ (e.g., driving ability). 
(iv) \textit{Asymmetry:} Trust is not necessarily symmetric; that is, the fact that node $u$ trusts node $v$ does not imply that $v$ also trusts $u$. 
(v) \textit{Conditional transitivity:} Trust is conditionally transitive or propagative, suggesting that it can be propagated from one node to another, creating a trust chain for two nodes that are not directly connected. For example, if node $u$ trusts node $k$, and $k$ trusts node $v$, then $u$ may also trust $v$ to some extent, provided that certain conditions are met, such as $u$'s trust standards~\cite{yan2008trust}.
(vi) \textit{Composability:} Due to the conditional transitivity of trust, a node may have several trust chains towards another. In this case, the overall trust between them is computed by aggregating trust levels from all chains (e.g., $u \rightarrow k_1 \rightarrow v$ and $u \rightarrow k_2 \rightarrow v$).

\textbf{Trust Evaluation.}
Trust evaluation is an important approach in cybersecurity, which quantifies a trustor's trust in a trustee based on relevant factors. Formally, given a trust graph, the \textit{trust evaluation problem} aims to evaluate (or predict) the trustworthiness of a trustor-trustee pair whose trust relationship is not explicitly represented in the graph.

A wide range of trust evaluation methods have been proposed, which can be broadly categorized into statistics-based, inference-based, and learning-based approaches~\cite{wang2022survey}. Statistics-based methods~\cite{massa2005controversial,chen2015trust} compute trust as a weighted sum of multiple trust-related factors. While simple and effective, they are highly sensitive to the choice of weights, making them difficult to generalize across different scenarios. Inference-based methods~\cite{chen2014trust} model trust with multiple dimensions and infer trust using predefined rules. For instance, OpinionWalk~\cite{liu2017opinionwalk} represents trust as a four-tuple and performs evaluation using discounting and combination operators derived from three-valued subjective logic~\cite{josang2006trust}. Similar to statistics-based methods, inference-based methods rely heavily on domain knowledge, which limits their generality. To address this limitation, learning-based methods have been developed, including those based on traditional neural networks~\cite{liu2019neuralwalk,wang2021c,jayasinghe2018machine} and Graph Neural Networks (GNNs)~\cite{wang2024trustguard,lin2020guardian,lin2021medley,jiang2022gatrust,huo2024trustgnn,wen2023dtrust}. Their primary advantage lies in the ability to automatically learn trust patterns or evaluation rules from large-scale data. NeuralWalk~\cite{liu2019neuralwalk}, for example, uses neural networks to learn trust propagation and aggregation but suffers from high computational overhead due to intensive matrix operations. In contrast, GNN-based methods utilize a ``message passing'' mechanism to propagate and aggregate trust information in a scalable manner~\cite{huo2024trustgnn,yu2023kgtrust}. They encode nodes and local neighborhood structures into low-dimensional embeddings, which can be used to evaluate trust at the node, pairwise, and group levels. Despite their promise for automated and accurate trust evaluation, learning-based approaches often require substantial labeled data, incur high training costs, lack full support for basic trust properties, and offer limited explainability. These limitations motivate the exploration of alternative approaches.

\subsection{Evaluation of LLMs}
Evaluating LLMs involves assessing their ability to meet predefined standards (knowledge-oriented) and perform specific tasks effectively (task-oriented)~\cite{wang2025the}.

\begin{figure*}[tb]
    \centering
    \includegraphics[width=\textwidth]{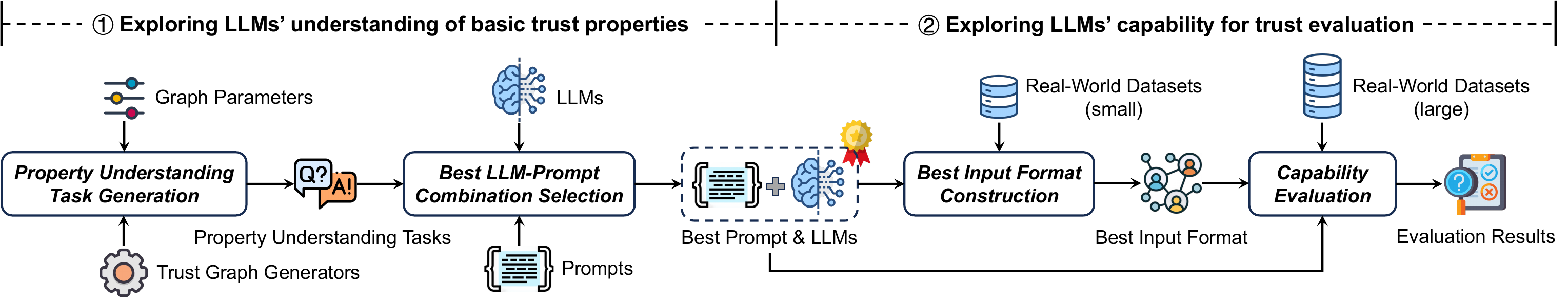}
    \vspace{-4mm}
    \caption{Overview of the LLM4Trust benchmark framework.}
    \label{framework}
    \vspace{-2mm}
\end{figure*}

\textbf{Knowledge-Oriented Assessment.}
This type of assessment examines LLMs' understanding of specialized knowledge domains, typically using multiple-choice questions derived from domain-specific materials. For instance, SecQA~\cite{liu2023secqa} includes approximately 200 questions to assess the knowledge of LLMs regarding security principles. CTIBench~\cite{alam2024ctibench} focuses on evaluating LLMs' comprehension of cyber threat intelligence concepts, including threat identification, detection strategies, mitigation techniques, etc. CyberMetric~\cite{tihanyi2024cybermetric} and SecEval~\cite{li2023seceval} assess expertise in areas such as penetration testing, cryptography, and network security. CSEBenchmark~\cite{wang2025the} offers a comprehensive evaluation framework containing 345 fine-grained knowledge points across seven cybersecurity subdomains. Beyond cybersecurity, knowledge-oriented assessments have also been conducted in the fields of ethics~\cite{rodionov2023evaluation}, medicine~\cite{zhang2025much}, and law~\cite{fei2023lawbench}.
Despite these efforts, trust-related concepts that differ fundamentally from the above cybersecurity knowledge have received limited attention. As described in Section~\ref{trust_basics}, trust is a complex concept with unique properties that are difficult to formalize and assess. Moreover, the absence of explicit annotations for these properties in existing datasets poses additional challenges for designing effective knowledge assessments in this domain.

\textbf{Task-Oriented Assessment.}
This type of assessment emphasizes the practical capabilities of LLMs in real-world applications. Recent studies have evaluated LLMs for the full lifecycle of vulnerability management~\cite{liu2024exploring}, vulnerability detection and analysis~\cite{li2025sv,ullah2024llms}, JavaScript deobfuscation~\cite{chen2025jsdeobsbench}, and Anomaly Detection (AD) tasks such as zero-shot AD, data augmentation, and AD model selection~\cite{yang2024ad}. In addition to cybersecurity tasks, LLMs have also been assessed on graph-related tasks~\cite{zhang2024llm4dyg,wang2025exploring,wang2023can}, news summarization~\cite{zhang2024benchmarking}, and financial analysis~\cite{xie2024finben}. While LLMs have been extensively evaluated across various domains, their capabilities for trust evaluation remain unexplored. Some graph-related benchmarks focus on general graph reasoning, such as shortest-path and connectivity reasoning. Although these capabilities are relevant to reasoning over trust graphs, they are not directly applicable to trust evaluation. In particular, real-world trust graphs exhibit unique characteristics~\cite{massa2009bowling,rossi2015network,kumar2016edge,kumar2018rev2}, including imbalanced trust distributions, multi-level trust relationships, and temporal dynamics, making it challenging to directly adapt existing benchmarks to trust evaluation.

To bridge these gaps, in this paper, we make the first attempt to explore the capability of LLMs for trust evaluation through both knowledge-oriented (\textbf{RQ1}) and task-oriented (\textbf{RQ2}) assessments.

\section{LLM4Trust Design}
In this section, we first overview the LLM4Trust benchmark and then describe its components in detail.

\subsection{Overview}
As shown in Fig.~\ref{framework}, LLM4Trust consists of two main phases: exploring LLMs' understanding of basic trust properties (\textbf{RQ1}) and exploring LLMs' capability for trust evaluation (\textbf{RQ2}). 
For \textbf{RQ1}, the key challenge lies in the inherent complexity of trust and the lack of explicit annotations for trust properties in existing datasets, which hinders the design of evaluation tasks. To address this challenge, we begin with the \textit{Property Understanding Task Generation} stage (Section~\ref{section_task}), where we design tasks targeting five basic trust properties. Specifically, we employ three data generators with varied parameters to construct trust graphs that exhibit diverse structural and statistical characteristics. Based on these graphs, we generate a number of question-answer pairs, referred to as property understanding tasks, to systematically assess LLMs' understanding of trust properties. In the \textit{Best LLM-Prompt Combination Selection} stage (Section~\ref{evaluation_rq1}), we input each question with a prompt into LLMs, and compare the generated responses against the ground-truth answers. This stage evaluates the accuracy of LLMs on the designed tasks under various prompt methods, allowing us to identify the most effective LLM-prompt combinations.

For \textbf{RQ2}, given the limited context windows of LLMs, directly feeding them complete real-world trust graphs is impractical. To address this challenge, we introduce the \textit{Best Input Format Construction} stage (Section~\ref{section_input}), where key subgraphs associated with trustor-trustee pairs are extracted under constraints on hop range and subgraph size. We tune these two parameters through experiments on randomly sampled small-scale datasets, aiming to reduce LLM query costs while identifying the best input format. In the \textit{Capability Evaluation} stage (Section~\ref{evaluation_rq2}), we apply the best-performing LLM-prompt combinations (identified in \textbf{RQ1}) and the best input format to large-scale real-world datasets, evaluating their performance in practical trust evaluation scenarios.

\subsection{Trust Graph Generators} \label{section_generators}
Due to the lack of datasets that explicitly model trust properties, we employ synthetic data generators to construct trust graphs with property-specific structures. These graphs enable us to design tasks for evaluating LLMs' understanding of trust properties. By default, we use the \textbf{Erdős-Rényi (ER) model}~\cite{erd6s1960evolution} to generate directed weighted graphs. Let $\mathcal{G} = (\mathcal{V}, \mathcal{E})$ denote a trust graph with node set $\mathcal{V}$ and edge set $\mathcal{E}$. We first generate a trust graph with the ER model $\mathcal{G} = ER(N, p)$, where $N$ is the number of nodes and $p$ is the probability that an edge exists between any pair of nodes. In this setup, $N$ controls the graph size, while $p$ determines the graph density. After generating $\mathcal{G}$, we assign attributes (e.g., timestamps $t$ and trust levels $w$) to edges based on the specific requirements of each property understanding task, as detailed in Section~\ref{section_task}. To enhance the diversity and representativeness of the generated graphs, we also employ the \textbf{Stochastic Block (SB) model}~\cite{holland1983stochastic} and the \textbf{Forest Fire (FF) model}~\cite{leskovec2007graph} for trust graph generation.

\subsection{Property Understanding Task Generation} \label{section_task}
\begin{figure}[tb]
    \centering
    \includegraphics[width=0.48\textwidth]{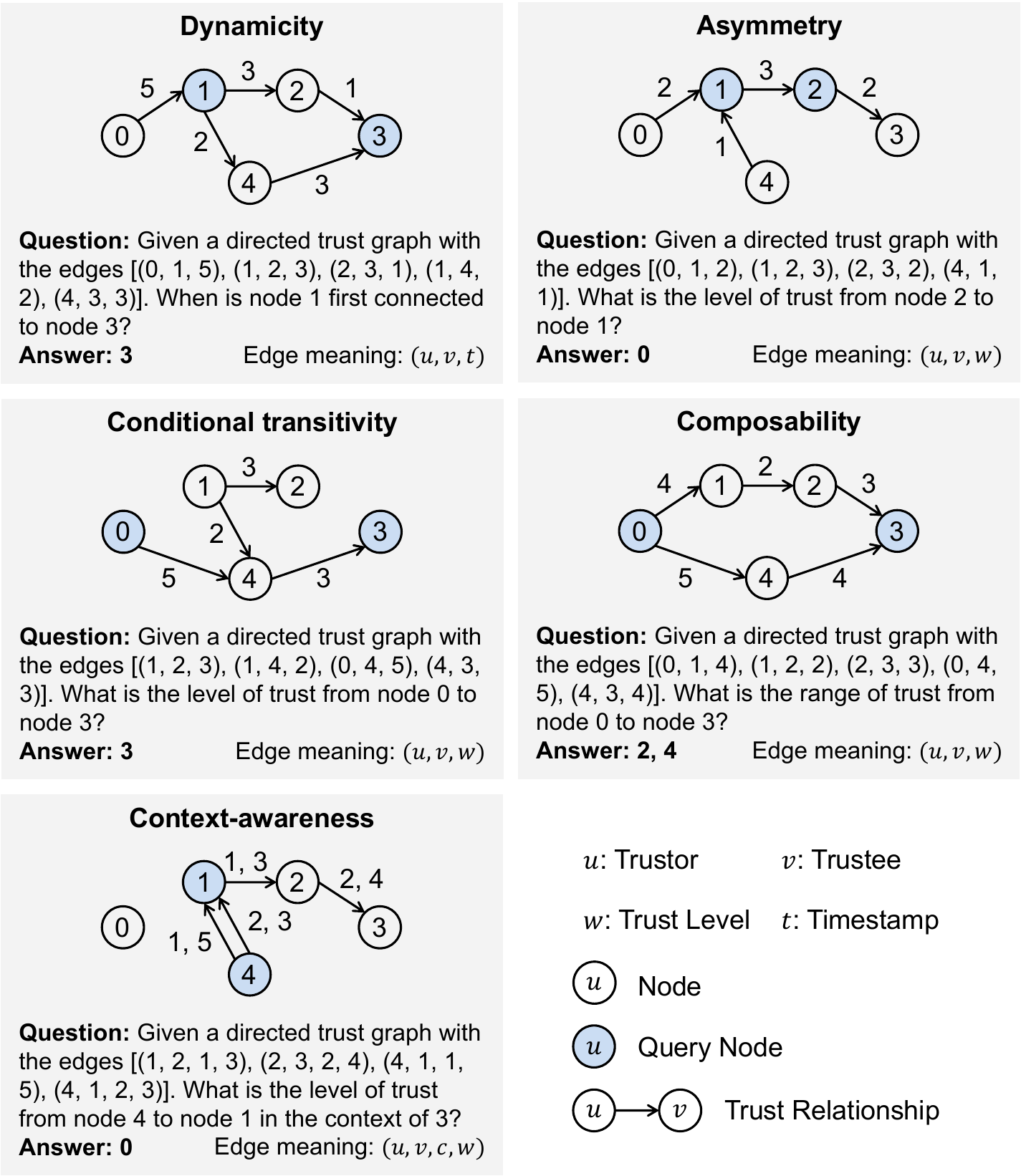}
    \caption{Overview of property understanding tasks.}
    \label{trust_tasks}
    \vspace{-1mm}
\end{figure}

Existing knowledge-oriented assessments typically rely on natural language representations and rarely cover trust-related concepts, making them unsuitable for assessing LLMs' understanding of trust properties. To address this gap, we design five property understanding tasks with graph representations. These tasks explicitly instantiate trust properties and enable their evaluation in controlled settings. Examples are illustrated in Fig.~\ref{trust_tasks}. Herein, we exclude subjectivity because (i) LLMs and other ML methods evaluate trust based on objectively available data, and (ii) interaction data, such as trust ratings from a trustor to a trustee, already encode the trustor's subjective perspective. For each task, we describe how to construct trust graphs and generate question-answer pairs based on trust theories and domain expertise. The pseudo-code for each task is provided in Appendix~\ref{appendix_algorithm}. For clarity, we denote the trustor as $u$ and the trustee as $v$ throughout the following descriptions.

\textbf{Understanding Dynamicity.}
This task evaluates whether LLMs can reason about temporal information in trust relationships. Given a node pair $(u,v)$, the goal is to identify the earliest time at which a trust chain is established from $u$ to $v$. Based on the graph $\mathcal{G}$ constructed in Section~\ref{section_generators}, we assign each edge a random timestamp $t$ drawn from a uniform distribution over $\{1, 2, \cdots, T\}$, where $T$ denotes the time span. A node pair $(u,v)$ with at least one trust chain (e.g., $u \xrightarrow{t} k_1 \xrightarrow{t} \cdots \xrightarrow{t} k_n \xrightarrow{t} v$) is randomly selected as a query to LLMs. The establishment time of a chain is defined as the maximum timestamp among its edges. If multiple chains exist, the minimum of their establishment times is taken as the ground-truth answer.

\textbf{Understanding Asymmetry.}
This task evaluates whether LLMs can recognize the asymmetric nature of trust. Given a node pair $(u,v)$, the goal is to infer the trust level from $v$ to $u$. Each edge in the graph $\mathcal{G}$ is assigned a random trust level $w$ drawn from a uniform distribution over $\{1,2,\cdots,W\}$, where $W$ denotes the number of trust levels. We randomly select a pair $(u,v)$ such that a direct trust relationship exists from $u$ to $v$, while no reverse trust relationship exists from $v$ to $u$, either directly or through trust chains. The query is then formulated as $(v,u)$, and the ground-truth answer is 0, indicating either distrust or the absence of trust in the reverse direction.

\textbf{Understanding Conditional Transitivity.}
This task assesses whether LLMs can understand the propagative nature of trust. Given a node pair $(u,v)$, the goal is to evaluate their trust relationship based on a trust chain of the form $u \rightarrow k_1 \rightarrow \cdots \rightarrow k_n \rightarrow v$, where each node places some trust in its successor. Similar to the asymmetry task, each edge in $\mathcal{G}$ is assigned a random trust level $w$. A pair $(u,v)$ connected by exactly one such chain is randomly selected as a query. Based on domain knowledge validated on real-world trust-related datasets~\cite{zhan2024enhancing}, the ground-truth answer is defined as the minimum trust level along the chain.

\textbf{Understanding Composability.}
This task examines whether LLMs can properly aggregate such information from different sources. Given a node pair $(u,v)$, the goal is to determine the range of their trust by considering all trust chains from $u$ to $v$. The graph $\mathcal{G}$ is initialized with random trust levels assigned to each edge, as in the previous tasks. A pair $(u,v)$ with at least two trust chains is randomly selected as a query. Based on domain knowledge, composable trust between a node pair can be defined in multiple ways, e.g., taking the maximum or minimum of the minimum trust levels across all trust chains~\cite{zhan2024enhancing}. Since no single aggregation rule is universally accepted, we define the ground-truth answer as a trust range between these two extremes, allowing multiple plausible aggregation outcomes.

\textbf{Understanding Context-Awareness.}
This task evaluates whether LLMs can distinguish trust relationships across different contexts. Given a node pair $(u,v)$, the goal is to assess their trust relationship under a specific context. To achieve this, each edge in the graph $\mathcal{G}$ is assigned a random context-specific trust level, forming a quadruple $(u,v,c,w)$. The quadruple denotes that trustor $u$ trusts trustee $v$ with level $w$ under context $c$. A pair $(u,v)$ with only direct trust relationships is selected, and LLMs are queried to evaluate their trust in a new, previously unseen context $c'$. The ground-truth answer is 0, indicating either distrust or the absence of trust in that new context.

The above synthetic tasks enable property-level trust reasoning evaluation in controlled settings, with ground truth defined according to trust reasoning rules derived from existing trust literature. These tasks are designed to reflect fundamental reasoning patterns involved in practical trust evaluation, including temporal evolution, trust propagation, multi-source trust aggregation, and directional and context-dependent trust reasoning. However, they do not fully capture the complexities of practice. Therefore, \textbf{RQ2} complements this evaluation by assessing LLMs on real-world trust networks with diverse and noisy trust relationships.

\subsection{Prompts}
Recent studies~\cite{shi2023large,wei2022chain,bai2024exploring} have shown that prompt design can significantly affect LLM performance. In LLM4Trust, we employ nine types of prompts, including six individual prompts and three combined prompts, to fully activate LLMs' capabilities for trust evaluation. These prompts support a thorough evaluation by covering three key dimensions: demonstration level, external knowledge, and reasoning guidance. 
Regarding demonstration level, we adopt \textbf{0-shot}, \textbf{1-shot}, and \textbf{few-shot} prompts to explore whether LLMs can perform trust-related tasks with no or minimal demonstration examples. By default, the few-shot setting uses three randomly selected examples, which yields strong performance (see Section~\ref{impacts}). 
To provide external knowledge, we adopt a \textbf{knowledge} prompt that incorporates domain knowledge (see Appendix~\ref{appendix_knowledge}) obtained from trust-related literature to support reasoning with expert insights. A \textbf{role} prompt is also adopted to assign a task-specific identity to LLMs, enhancing contextual alignment and task comprehension. 
For reasoning guidance, we include a \textbf{Chain-of-Thought (CoT)} prompt to encourage step-by-step reasoning for structured and accurate inference. Additionally, we design three combined prompts, namely \textbf{CoT+knowledge+role}, \textbf{few-shot+knowledge+role}, and \textbf{few-shot+CoT+knowledge+role}, to exploit the complementary strengths of individual prompts.

In addition to these prompts, a complete prompt template also includes instructions for the trust graph, task, and answer format, as well as a specific question that defines the trust evaluation scenario. We provide the full prompt templates for all prompt methods in Appendix~\ref{appendix_prompt}.

\subsection{LLMs} \label{section_llm}
We evaluate seven general-purpose LLMs: \textbf{GPT-3.5}, \textbf{GPT-4o}, \textbf{DeepSeek-V3}~\cite{liu2024deepseek}, \textbf{Qwen-2.5-Max}~\cite{qwen25}, \textbf{Llama-4-Scout}, \textbf{Llama-4-Maverick}, and \textbf{Claude-3.7-Sonnet}. These models are selected to ensure diversity in architectures, parameter sizes, context window sizes, and knowledge cutoffs. While they exhibit some reasoning capability, we distinguish them from reasoning-oriented models (e.g., GPT-5) that are explicitly designed to support complex multi-step reasoning. We therefore include \textbf{GPT-5} for comparison. Detailed specifications of all evaluated LLMs are summarized in Table~\ref{llm_statistics}. All of them are accessed via their official APIs. Compared to locally deployed LLMs, API-based access offers two key advantages: (i) It ensures stable access to the latest official model versions and updates, avoiding inconsistencies caused by varying local deployment environments, hardware configurations, or unofficial model weights. (ii) It enables evaluation under realistic usage scenarios, reflecting the common practice in which users interact with LLMs through APIs. When invoking the APIs, two critical parameters should be considered, namely \textit{temperature} and \textit{top\_p}. The \textit{temperature} parameter controls the ``creativity'' or randomness of the responses generated by LLMs, while \textit{top\_p} controls response diversity by limiting sampling to the most likely tokens. To enhance response consistency and reduce sampling variance across multiple queries, we follow the guidance of Liu \textit{et al.}~\cite{liu2024exploring} by setting the \textit{temperature} to 0 and \textit{top\_p} to 1.0. A detailed analysis of their effects is provided in Appendix~\ref{appendix_temperature}.

\begin{table}[tbp]
\centering
\caption{Comparison of selected LLMs.}
\label{llm_statistics}
\renewcommand{\arraystretch}{1.1}
\resizebox{\linewidth}{!}{
\begin{tabular}{lcccc}
\toprule
\textbf{Model API} & \textbf{Base Model} & \textbf{\# Params} & \textbf{\makecell[c]{Context Size}} & \textbf{\makecell[c]{Knowledge \\Cutoff}} \\
\midrule
gpt-3.5-turbo & GPT-3.5 & -- & 16,385 & 09/2021 \\
gpt-4o & GPT-4o & -- & 128k & 10/2023 \\
deepseek-chat & DeepSeek-V3 & 671B & 128k & 07/2024 \\
qwen-max & Qwen-2.5-Max & -- & 32k & 10/2023 \\
llama-4-scout-17b-16e-instruct & Llama-4-Scout & 109B & 10M & 08/2024 \\
llama-4-maverick-17b-128e-instruct & Llama-4-Maverick & 400B & 1M & 08/2024 \\
claude-3-7-sonnet-20250219 & Claude-3.7-Sonnet & -- & 200k & 10/2024 \\
gpt-5-2025-08-07 & GPT-5 & -- & 400k & 09/2024 \\
\bottomrule
\end{tabular}
}
\vspace{-1mm}
\end{table}

\subsection{Real-World Datasets}
We utilize five real-world trust networks collected from different domains. \textbf{Advogato}~\cite{massa2009bowling} comes from an online social network for open-source software developers, where a trust relationship indicates that a trustor trusts a trustee's software development skills. \textbf{Pretty-Good-Privacy (PGP)}~\cite{rossi2015network} is obtained from a public certification network, where trust reflects one user's attestation of another's trustworthiness. \textbf{Epinions}~\cite{tang2015trust} is obtained from consumer review sites where users build a ``Web of Trust'' by adding reviewers they find valuable. The above datasets are static with either binary (trust or distrust) or four trust levels. \textbf{Bitcoin-OTC} and \textbf{Bitcoin-Alpha}~\cite{kumar2016edge,kumar2018rev2} are collected from Bitcoin trading platforms. Since Bitcoin users are anonymous, evaluating their trust is crucial for preventing potential fraud. In these two datasets, trust relationships are timestamped, forming dynamic trust graphs. Table~\ref{dataset_statistics} presents the statistics of the five datasets. By selecting datasets that cover both static and dynamic graphs, as well as binary and multi-level trust relationships, we ensure broad coverage of real-world trust scenarios in our evaluation.

\begin{table}[tbp]
\centering
\caption{Statistics of real-world datasets.}
\label{dataset_statistics}
\renewcommand{\arraystretch}{1.1}
\resizebox{\linewidth}{!}{
\begin{tabular}{lccccccc}
\toprule
\textbf{Dataset} & \textbf{\# Nodes} & \textbf{\# Edges} & \textbf{\makecell[c]{Avg. \\Degree}} & \textbf{\makecell[c]{\# Trust \\Levels}} & \textbf{\makecell[c]{\makecell[c]{Trust \\Ratio}}} & \textbf{\makecell[c]{\makecell[c]{Graph \\Type}}} & \textbf{Domain} \\
\midrule
Advogato & 5,280 & 54,382 & 20.6 & 4 & 32.4\% & static & social \\
PGP & 37,841 & 317,081 & 16.8 & 4 & 30.3\% & static & security \\
Epinions & 18,098 & 711,508 & 78.6 & 2 & 50.0\% & static & e-commerce \\
Bitcoin-OTC & 5,881 & 35,592 & 12.1 & 2 & 90.0\% & dynamic & financial \\
Bitcoin-Alpha & 3,783 & 24,186 & 12.8 & 2 & 93.7\% & dynamic & financial \\
\bottomrule
\end{tabular}
}
\vspace{-2mm}
\end{table}

\subsection{Best Input Format Construction} \label{section_input}
Real-world trust networks are often too large to be directly fed into LLMs due to their limited context windows. To address this challenge, we extract a subgraph that captures the most relevant trust information for each trustor-trustee pair and use it as the LLM input. Two key factors guide the subgraph construction: the hop range, which refers to the maximum number of hops between the trustor and the trustee, and the subgraph size, measured by the number of edges. To efficiently determine the optimal parameter settings (i.e., the best input format), we conduct a parameter study on randomly sampled real-world data using the best-performing LLM-prompt combinations from \textbf{RQ1}. We follow a structural principle that prioritizes shorter-range neighbors (e.g., one-hop neighbors are preferred over two-hop ones), as they capture more direct and relevant trust information than distant neighbors~\cite{wang2024trustguard}. While this strategy works well for static trust graphs, it faces additional challenges in dynamic trust graphs.

A key characteristic of dynamic graphs is that interactions are timestamped. Accordingly, we adopt a temporal principle for subgraph construction: recent interactions are prioritized, as they better reflect current trust states than outdated ones~\cite{lin2021medley,wang2025cat}. Moreover, nodes may appear (start interacting) or become inactive over time. Thus, when chronologically dividing a dynamic dataset into training and testing sets, we define unobserved nodes as those that appear only in the testing phase, while observed nodes are present in both phases. Based on whether the trustor and trustee are observed or unobserved, we identify four trust evaluation cases:
\begin{itemize}[leftmargin=*]
    \item \textbf{Case 1:} Both trustors and trustees are observed. We extract subgraphs related to trustor-trustee pairs following the proposed structural and temporal principles.
    \item \textbf{Case 2:} Trustors are observed while trustees are unobserved. We extract a 2-hop out-degree subgraph centered on the trustor, as prior studies~\cite{huo2024trustgnn} have shown that 2-hop neighbors are sufficient for trust propagation and aggregation. This subgraph captures the trustor's subjective trust preferences, i.e., the degree to which it tends to trust others, which are essential for evaluating its trust towards unobserved nodes.
    \item \textbf{Case 3:} Trustors are unobserved while trustees are observed. Unlike Case 2, we extract a 2-hop in-degree subgraph centered on the trustee, given the effectiveness of 2-hop neighbors in capturing trust information~\cite{huo2024trustgnn}. This subgraph reflects the trustee's objective trustworthiness, i.e., the degree to which it tends to be trusted by others.
    \item \textbf{Case 4:} Both trustors and trustees are unobserved, making this case particularly challenging due to the absence of trust-related data for either party. To address this, we construct a subgraph using the most recent interactions in the dynamic trust graph. The intuition is that recent trust trends among observed pairs can serve as a proxy for inferring trust between unobserved pairs.
\end{itemize}

\section{Evaluation on LLMs' Understanding of Basic Trust Properties (RQ1)} \label{evaluation_rq1}
In this section, we present a fine-grained evaluation of LLMs' understanding of basic trust properties across different models, prompt methods, trust graph configurations, etc.

\subsection{Evaluation Setup}
\textbf{Evaluation Metric.}
We employ accuracy to evaluate LLM performance on property understanding tasks, as it provides an intuitive measure of correctness. Accuracy is computed as the ratio of correctly answered questions to the total number of questions. Higher accuracy indicates a better understanding of trust properties.

\textbf{Baselines.}
To verify whether LLMs can understand basic trust properties instead of producing random answers, we introduce a baseline that randomly selects answers from the candidate answer space. Since each property understanding task has a different answer space, the baseline accuracy varies accordingly. For understanding dynamicity, the baseline accuracy is $1/T$, where $T$ is the time span. For understanding conditional transitivity, it is $1/W$, where $W$ denotes the number of trust levels. For understanding asymmetry and context-awareness, the baseline accuracy is $1/(W+1)$ since both tasks consider the possibility of distrust or the absence of trust. For understanding composability, the baseline accuracy is $2/[W(W+1)]$, as the task requires outputting trust ranges. Beyond random guessing, we also include two heuristic baselines for conditional transitivity and composability: Mean, which aggregates trust by averaging trust levels along a trust chain, and Decay, which models trust decay over long chains.

\textbf{Implementation Details.}
By default, we generate trust graphs using the ER model with $N=5$ and $p=0.3$. This setting is sufficient to construct valid trust graphs for property understanding tasks, while keeping the generated graphs compact and the modeled trust properties clearly identifiable. For understanding dynamicity, the time span $T$ is set to 5. The number of trust levels $W$ is set to 5, except in the context-awareness task, where it is increased to 9 to ensure diverse trust levels across contexts. To evaluate LLMs under different graph configurations, we vary the above parameters (i.e., $N$, $p$, $T$, and $W$) and trust graph generators, and analyze their effects in Section~\ref{impacts}. For each task and setting, we randomly generate 1,000 question-answer pairs for evaluation. Each experiment is repeated with two random seeds, resulting in 10,000 pairs in total (5 tasks $\times$ 1,000 pairs $\times$ 2 seeds). All reported results are averaged over the two runs.

\subsection{Main Results}

\begin{table*}[htbp]
\centering
\scriptsize
\caption{Accuracy of LLMs' understanding of trust properties with different prompts.}
\label{main_results_rq1}
\newcolumntype{C}[1]{>{\centering\arraybackslash}p{1.1cm}}
\resizebox{\linewidth}{!}{
\begin{threeparttable}
\begin{tabular}{ll*{9}{C{1.1cm}}}
\toprule
\textbf{Task} & \textbf{Model} & \textbf{0-shot} & \textbf{1-shot} & \textbf{Few-shot} & \textbf{Know} & \textbf{Role} & \textbf{CoT} & \makecell{\textbf{CKR}} & \makecell{\textbf{FKR}} & \makecell{\textbf{FCKR}} \\
\midrule

\multirow{8}{*}{\textbf{Dynamicity}} 
& Random & \multicolumn{9}{c}{0.20} \\
\cline{2-11}
& GPT-3.5 & 0.75 & 0.75 & 0.66 & 0.78 & 0.74 & 0.72 & 0.70 & 0.77 & 0.63 \\
& GPT-4o & 0.85 & 0.96 & 0.96 & 0.84 & 0.81 & 0.85 & 0.83 & 0.98 & 0.98 \\
& DeepSeek-V3 & 0.82 & 0.91 & 0.95 & 0.80 & 0.79 & 0.85 & 0.81 & 0.92 & 0.94 \\
& Qwen-2.5-Max & 0.84 & 0.95 & 0.95 & 0.85 & 0.84 & 0.88 & 0.82 & 0.95 & 0.93 \\
& Llama-4-Scout & 0.82 & 0.92 & 0.86 & 0.81 & 0.85 & 0.82 & 0.81 & 0.90 & 0.91 \\
& Llama-4-Maverick & 0.94 & 0.98 & 0.97 & 0.92 & 0.94 & 0.96 & 0.94 & 0.98 & 0.98 \\
& Claude-3.7-Sonnet & 0.91 & 0.97 & 0.98 & 0.89 & 0.88 & 0.92 & 0.86 & \textbf{1.00} & 0.99 \\
\midrule

\multirow{8}{*}{\textbf{Asymmetry}} 
& Random & \multicolumn{9}{c}{0.17} \\
\cline{2-11}
& GPT-3.5 & 0.19 & 0.20 & 0.58 & 0.46 & 0.16 & 0.09 & 0.22 & 0.37 & 0.29 \\
& GPT-4o & \textbf{1.00} & \textbf{1.00} & \textbf{1.00} & \textbf{1.00} & \textbf{1.00} & \textbf{1.00} & \textbf{1.00} & \textbf{1.00} & \textbf{1.00} \\
& DeepSeek-V3 & \textbf{1.00} & \textbf{1.00} & \textbf{1.00} & \textbf{1.00} & \textbf{1.00} & \textbf{1.00} & \textbf{1.00} & \textbf{1.00} & \textbf{1.00} \\
& Qwen-2.5-Max & \textbf{1.00} & \textbf{1.00} & \textbf{1.00} & \textbf{1.00} & 0.98 & 0.97 & \textbf{1.00} & 0.99 & \textbf{1.00} \\
& Llama-4-Scout & 0.91 & 0.98 & 0.99 & 0.98 & 0.91 & 0.87 & 0.99 & \textbf{1.00} & \textbf{1.00} \\
& Llama-4-Maverick & 0.98 & \textbf{1.00} & \textbf{1.00} & 0.99 & 0.96 & 0.98 & 0.98 & \textbf{1.00} & \textbf{1.00} \\
& Claude-3.7-Sonnet & \textbf{1.00} & \textbf{1.00} & \textbf{1.00} & \textbf{1.00} & \textbf{1.00} & \textbf{1.00} & \textbf{1.00} & \textbf{1.00} & \textbf{1.00} \\
\midrule

\multirow{8}{*}{\textbf{\makecell{Conditional\\Transitivity}}} 
& Random / Mean / Decay & \multicolumn{9}{c}{0.20 / 0.24 / 0.79} \\
\cline{2-11}
& GPT-3.5 & 0.19 & 0.35 & 0.32 & 0.50 & 0.21 & 0.20 & 0.65 & 0.63 & 0.69 \\
& GPT-4o & 0.16 & 0.81 & 0.98 & 0.98 & 0.29 & 0.23 & 0.98 & 0.96 & 0.97 \\
& DeepSeek-V3 & 0.16 & 0.38 & 0.88 & \textbf{1.00} & 0.26 & 0.26 & 0.99 & \textbf{1.00} & \textbf{1.00} \\
& Qwen-2.5-Max & 0.16 & 0.36 & 0.67 & 0.95 & 0.28 & 0.45 & 0.95 & 0.91 & 0.96 \\
& Llama-4-Scout & 0.57 & 0.80 & 0.89 & 0.93 & 0.49 & 0.63 & 0.94 & 0.88 & 0.89 \\
& Llama-4-Maverick & 0.17 & 0.94 & 0.98 & \textbf{1.00} & 0.18 & 0.31 & \textbf{1.00} & \textbf{1.00} & 0.99 \\
& Claude-3.7-Sonnet & 0.05 & 0.39 & 0.99 & \textbf{1.00} & 0.07 & 0.04 & \textbf{1.00} & \textbf{1.00} & \textbf{1.00} \\
\midrule

\multirow{8}{*}{\textbf{Composability}} 
& Random / Mean / Decay & \multicolumn{9}{c}{0.07 / 0.03 / 0.61} \\
\cline{2-11}
& GPT-3.5 & 0.11 & 0.08 & 0.06 & 0.26 & 0.05 & 0.06 & 0.59 & 0.43 & 0.56 \\
& GPT-4o & 0.15 & 0.07 & 0.06 & 0.82 & 0.32 & 0.25 & 0.83 & 0.79 & 0.81 \\
& DeepSeek-V3 & 0.52 & 0.69 & 0.75 & 0.94 & 0.74 & 0.61 & 0.95 & 0.90 & 0.92 \\
& Qwen-2.5-Max & 0.56 & 0.62 & 0.47 & 0.77 & 0.46 & 0.63 & 0.68 & 0.53 & 0.68 \\
& Llama-4-Scout & 0.70 & 0.67 & 0.68 & 0.85 & 0.76 & 0.73 & 0.79 & 0.80 & 0.78 \\
& Llama-4-Maverick & 0.73 & 0.87 & 0.90 & 0.96 & 0.72 & 0.64 & \textbf{0.97} & \textbf{0.97} & 0.94 \\
& Claude-3.7-Sonnet & 0.71 & 0.89 & 0.93 & 0.94 & 0.81 & 0.74 & \textbf{0.97} & 0.96 & 0.96 \\
\midrule

\multirow{8}{*}{\textbf{\makecell{Context-\\Awareness}}} 
& Random & \multicolumn{9}{c}{0.10} \\
\cline{2-11}
& GPT-3.5 & 0.35 & 0.33 & 0.59 & 0.72 & 0.23 & 0.24 & 0.21 & 0.70 & 0.48 \\
& GPT-4o & \textbf{1.00} & \textbf{1.00} & \textbf{1.00} & \textbf{1.00} & \textbf{1.00} & 0.99 & \textbf{1.00} & \textbf{1.00} & \textbf{1.00} \\
& DeepSeek-V3 & \textbf{1.00} & \textbf{1.00} & \textbf{1.00} & \textbf{1.00} & \textbf{1.00} & \textbf{1.00} & \textbf{1.00} & \textbf{1.00} & \textbf{1.00} \\
& Qwen-2.5-Max & 0.94 & 0.94 & 0.98 & \textbf{1.00} & 0.91 & 0.97 & \textbf{1.00} & \textbf{1.00} & \textbf{1.00} \\
& Llama-4-Scout & 0.94 & 0.86 & 0.91 & \textbf{1.00} & 0.93 & 0.93 & 0.98 & \textbf{1.00} & \textbf{1.00} \\
& Llama-4-Maverick & 0.89 & 0.95 & 0.99 & 0.99 & 0.81 & 0.89 & 0.98 & \textbf{1.00} & \textbf{1.00} \\
& Claude-3.7-Sonnet & \textbf{1.00} & \textbf{1.00} & \textbf{1.00} & \textbf{1.00} & \textbf{1.00} & \textbf{1.00} & \textbf{1.00} & \textbf{1.00} & \textbf{1.00} \\
\bottomrule
\end{tabular}

\begin{tablenotes}[para,flushleft]
\item[] The best result for each task is highlighted in \textbf{bold}. Know: Domain knowledge; CKR: CoT+Knowledge+Role; FKR: Few-shot+Knowledge+Role; FCKR: Few-shot+CoT+Knowledge+Role.
\end{tablenotes}
\end{threeparttable}
  
}
\vspace{-4mm}
\end{table*}

Table~\ref{main_results_rq1} presents the accuracy of LLMs' understanding of trust properties under different prompts. We observe that the best-performing LLM-prompt combinations achieve nearly 100\% accuracy across all tasks, with relative improvements ranging from 27\% to 3133\% compared to the baselines. These results demonstrate that LLMs can effectively understand the questions in each task and infer trust levels based on specific properties, rather than making random guesses or relying on simple heuristics. Notably, some LLMs perform well even in the 0-shot setting, suggesting that trust-related concepts may have been included during pretraining. To understand what affects LLM accuracy, we identify three key factors:

\begin{figure}[tb]
    \centering
    \includegraphics[width=0.48\textwidth]{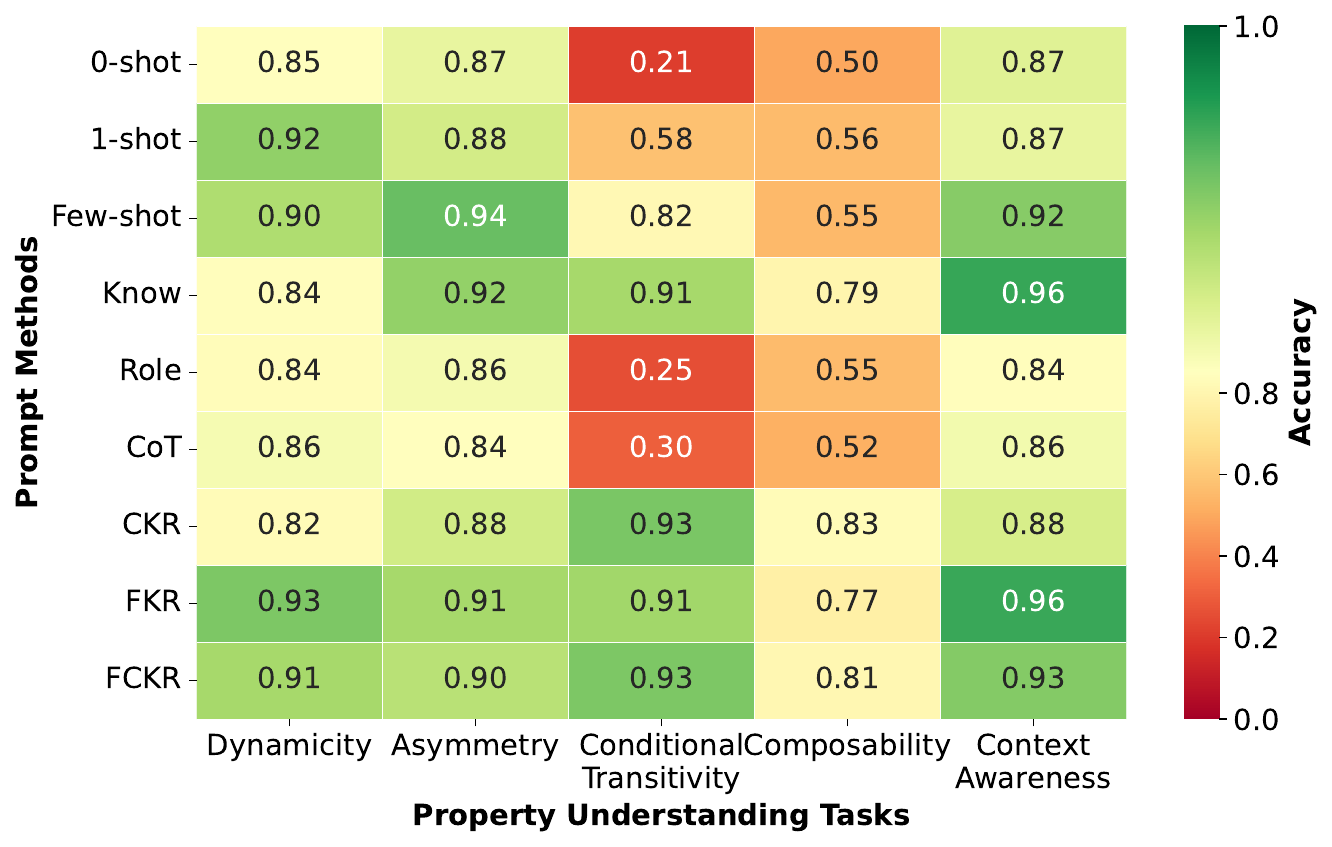}
    \vspace{-1mm}
    \caption{Average accuracy of prompt methods across property understanding tasks.}
    \label{prompt_task}
    \vspace{-4mm}
\end{figure}

\textbf{Prompt Design.} As shown in Fig.~\ref{prompt_task}, performance varies substantially across prompt methods. Among individual prompts, ``few-shot'' and ``knowledge'' perform best. This is because demonstration examples illustrate how to reason from graph structures to trust levels, while domain knowledge provides explicit trust reasoning principles. Leveraging these complementary strengths, combined prompts, particularly ``few-shot+knowledge+role'', outperform all individual prompts. These results highlight the importance of prompt design for high-quality trust evaluation.

\textbf{Task Difficulty.} We can observe from Fig.~\ref{prompt_task} that understanding conditional transitivity and composability is much more challenging than understanding other trust properties. This difficulty arises because conditional transitivity corresponds to trust propagation rules, while composability involves trust aggregation rules. Both types of rules are fundamental to trust evaluation~\cite{liu2017opinionwalk}, but they are inherently complex and lack unique definitions in the literature, which may confuse LLMs.

\begin{figure}[tb]
    \centering
    \includegraphics[width=0.48\textwidth]{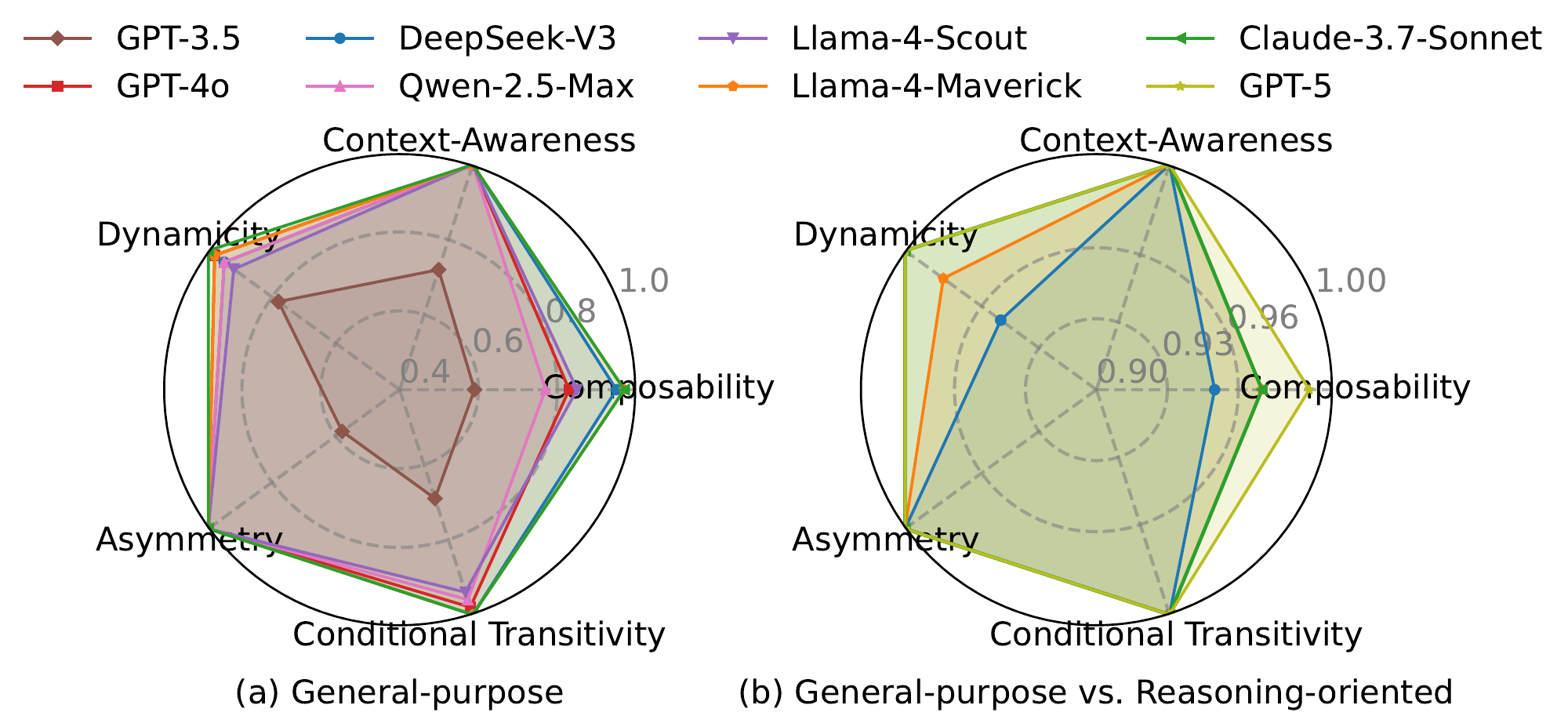}
    \vspace{-1mm}
    \caption{Maximum accuracy of each LLM across property understanding tasks.}
    \label{radar}
    \vspace{-3mm}
\end{figure}

\textbf{Model Capability.} Table~\ref{main_results_rq1} and Fig.~\ref{radar}(a) show that smaller LLMs with earlier knowledge cutoffs (e.g., GPT-3.5) perform poorly, even falling below baselines on a few tasks. In contrast, larger and more recent models, including Claude-3.7-Sonnet, Llama-4-Maverick, and DeepSeek-V3, show clear advantages. These advantages stem from not only access to more up-to-date training data, but also improvements in model architectures that enhance their reasoning capabilities. Among them, Claude-3.7-Sonnet achieves the best performance, reaching 100\% accuracy on four tasks and 97\% accuracy on the composability understanding task with appropriate prompts.

\begin{tcolorbox}[
    colback=white,
    colframe=black,
    boxrule=1pt,
    arc=0pt,
    left=3pt,
    right=3pt,
    top=3pt,
    bottom=3pt,
    boxsep=0pt
]
\textbf{Finding 1.} LLMs exhibit strong understanding of basic trust properties, achieving near-perfect accuracy with appropriate prompts. Their performance is influenced by three key factors: prompt design (combined prompts outperform individual ones), task difficulty (understanding composability is the most challenging task), and model capability (larger models with more recent knowledge cutoffs perform better).
\end{tcolorbox}

\subsection{Impact of Trust Graph Configurations, Shot Numbers, and Model Types} \label{impacts}
Unless otherwise specified, all experiments in this subsection use the best-performing LLM-prompt combination (i.e., Claude-3.7-Sonnet with ``few-shot+knowledge+role'').

\begin{figure}[tb]
    \centering
    \includegraphics[width=0.47\textwidth]{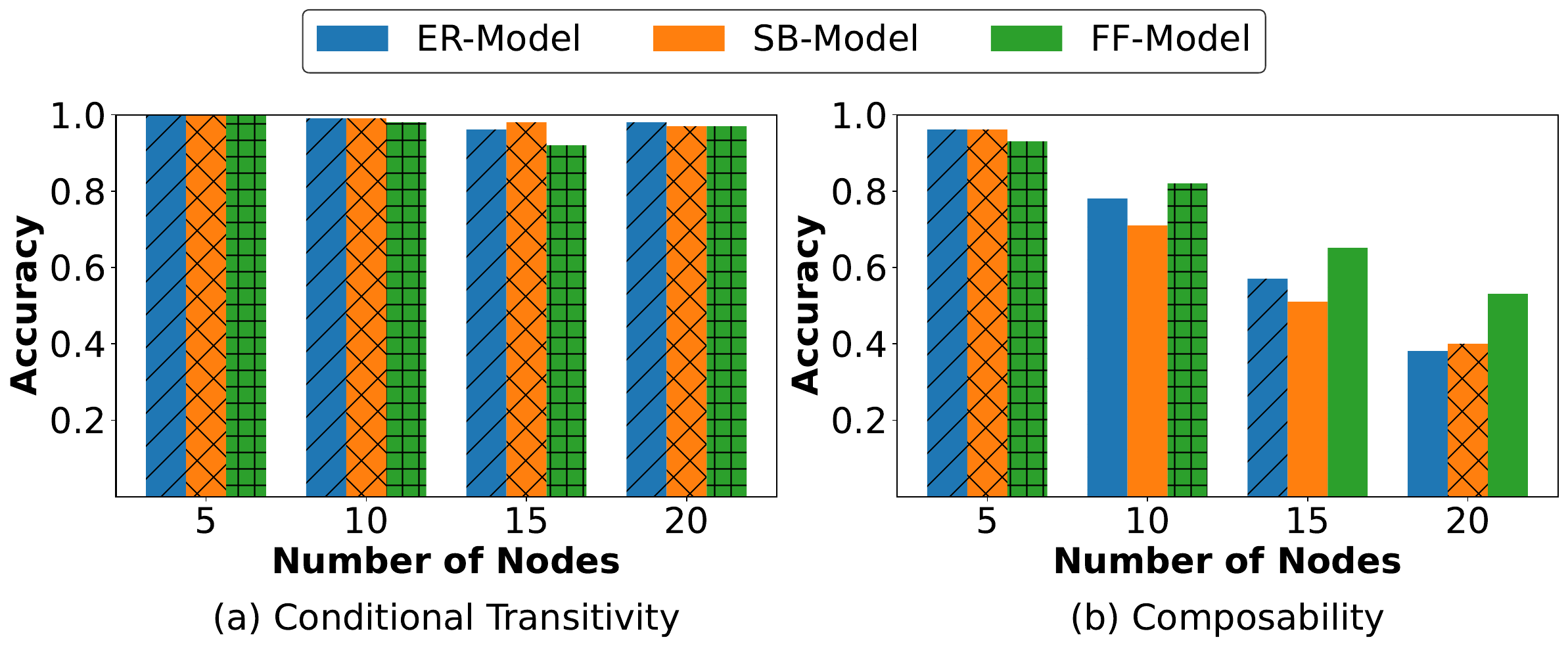}
    \vspace{-1mm}
    \caption{Impact of trust graph generators and graph sizes.}
    \label{generators}
    \vspace{-4mm}
\end{figure}

\textbf{Impact of Trust Graph Generators and Graph Sizes.}
We investigate how graph topology (induced by different trust graph generators) and graph size affect LLMs' understanding of trust properties. Our analysis focuses on conditional transitivity and composability, the two most challenging tasks. 
As shown in Fig.~\ref{generators}(a), LLMs achieve over 90\% accuracy across all graph sizes and generator types, indicating that they are robust to structural variations in the conditional transitivity task. In contrast, for composability shown in Fig.~\ref{generators}(b), performance degrades substantially and exhibits high variance as graph size increases. This trend reveals that while LLMs can reliably reason over small trust graphs, their reasoning capabilities do not scale well to large graphs. These results highlight the importance of controlling input graph size when applying LLMs to real-world trust evaluation.

\begin{figure}[tb]
    \centering
    \includegraphics[width=0.47\textwidth]{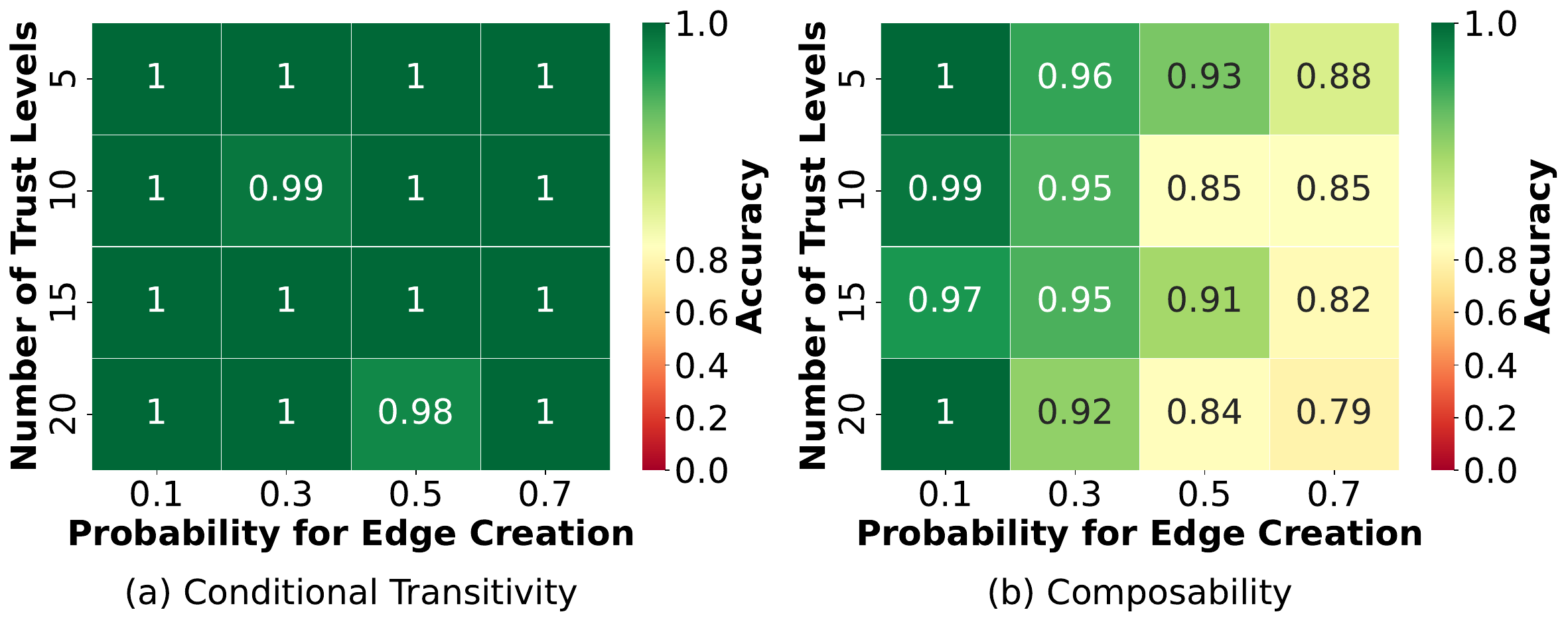}
    \vspace{-1mm}
    \caption{Impact of graph density and trust level range.}
    \label{graph_statistics}
    \vspace{-3mm}
\end{figure}

\textbf{Impact of Other Graph Parameters.}
We explore how graph parameters, including the number of trust levels, graph density, and time span, affect LLM accuracy. Specifically, we use the two most challenging tasks to study the first two parameters, as they require multi-hop reasoning and may be sensitive to trust level range and structural density. As shown in Fig.~\ref{graph_statistics}, the LLM consistently achieves over 98\% accuracy across different settings for the conditional transitivity task, suggesting that it can support multi-level trust evaluation and remains stable under varying graph densities. In contrast, the composability task is relatively sensitive to these parameters. As the edge creation probability increases, the graph contains more possible trust propagation paths, making it harder for the LLM to identify and combine the relevant relationships correctly. These results indicate that learning trust propagation rules is largely unaffected by graph density and trust level range, whereas composability reasoning becomes increasingly challenging in denser graphs with more trust levels.

\begin{figure}[tb]
    \begin{minipage}[t]{0.47\linewidth}
        \centering
        \includegraphics[width=\linewidth]{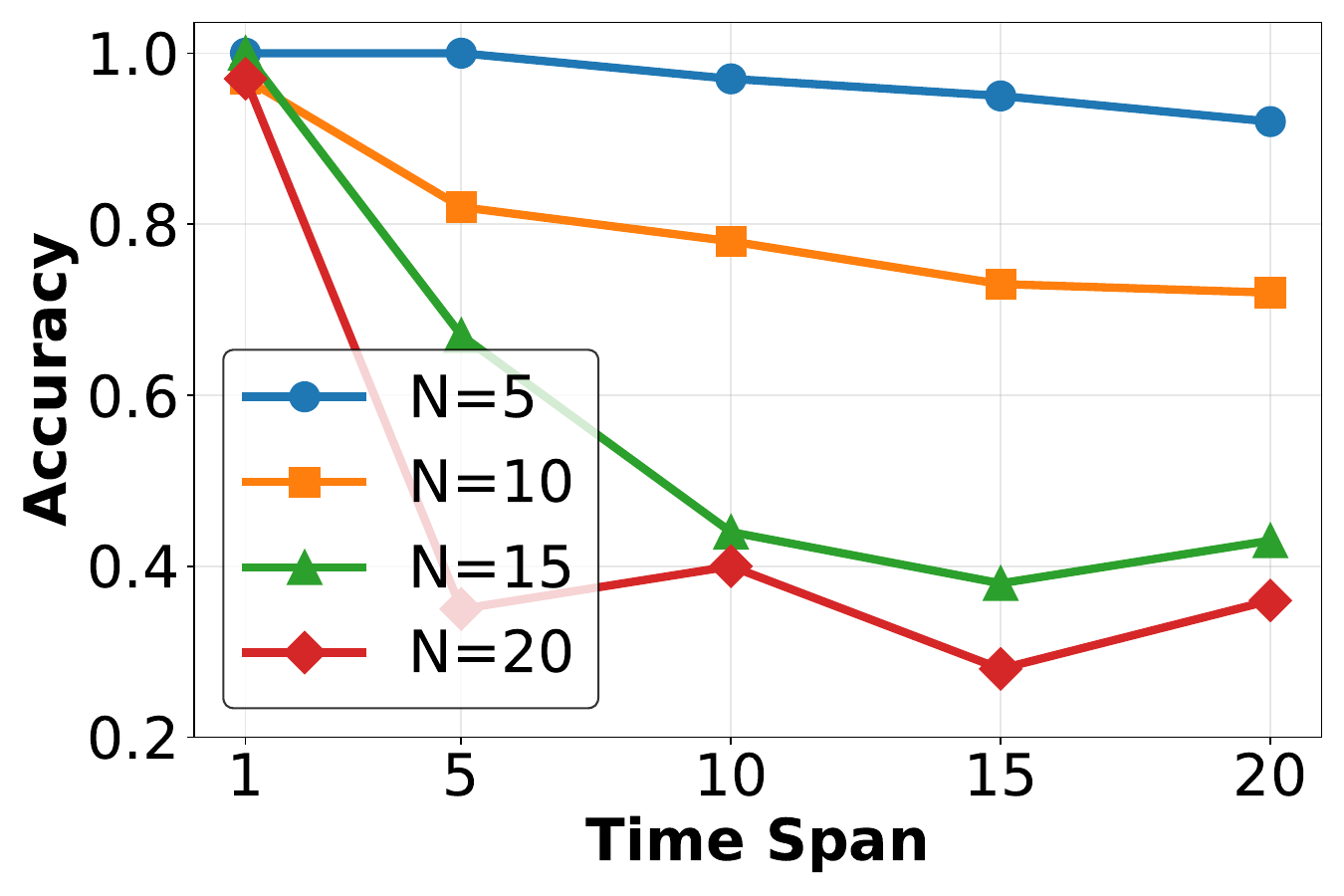}
        \vspace{-5mm}
        \caption{Impact of time span across different graph sizes.}
        \label{time}
    \end{minipage}
    \hfill
    \begin{minipage}[t]{0.47\linewidth}
        \centering
        \includegraphics[width=\linewidth]{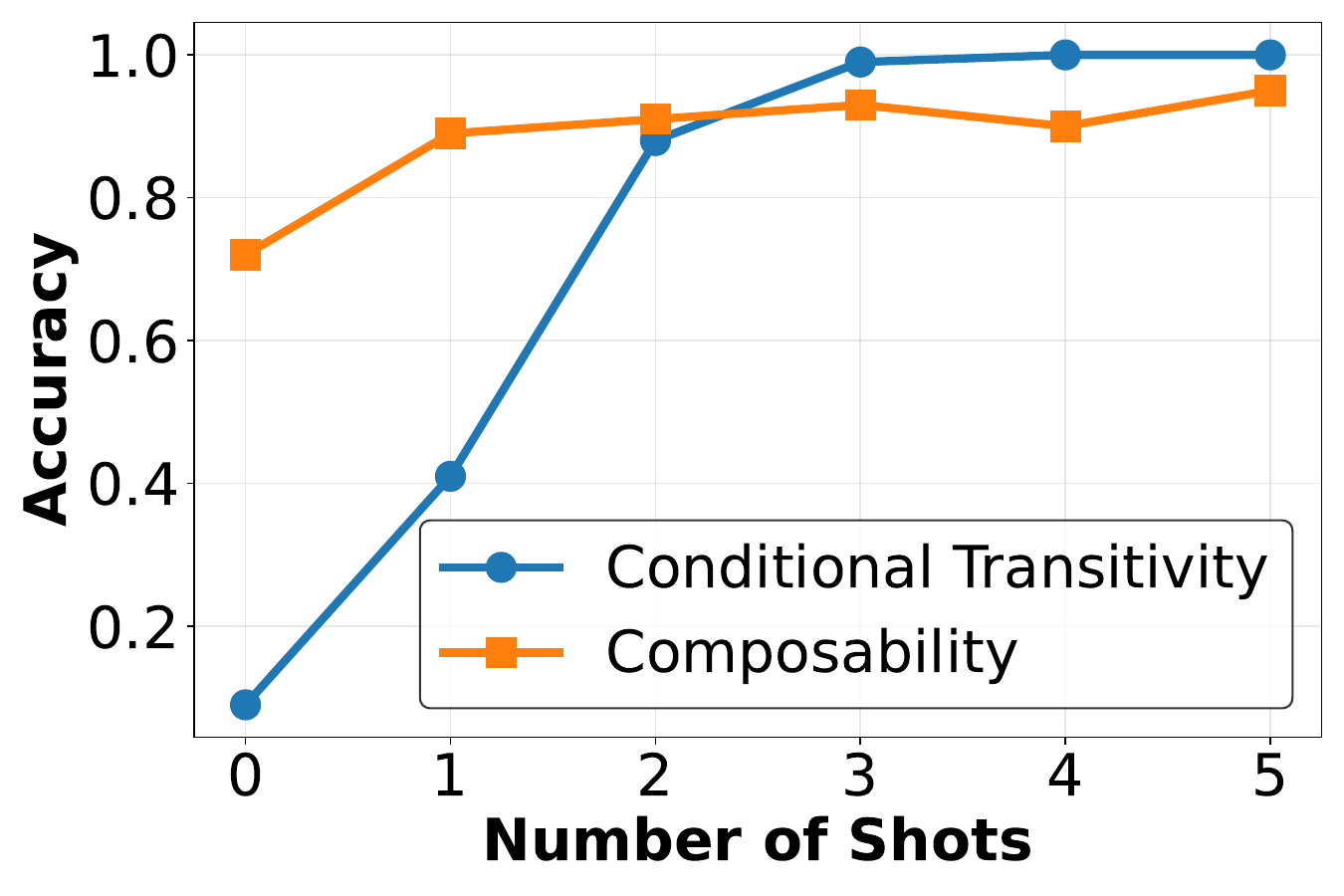}
        \vspace{-5mm}
        \caption{Impact of different shot numbers.}
        \label{few_shots}
    \end{minipage}
\vspace{-4mm}
\end{figure}

To assess the impact of time span, we use the dynamicity task, which explicitly models temporal changes in trust relationships. Fig.~\ref{time} shows that LLM accuracy declines as time span increases, indicating that understanding trust dynamicity is highly related to temporal complexity: the wider the time span, the more difficult the task becomes. This difficulty is further amplified in large graphs, highlighting the challenges of applying LLMs to large-scale dynamic trust graphs.

\textbf{Impact of Shot Numbers.}
We study how the number of shots (i.e., demonstration examples) affects LLM accuracy. Similarly, we focus on the two most challenging tasks.
As shown in Fig.~\ref{few_shots}, the LLM exhibits a clear upward trend in accuracy on both tasks as the number of shots increases, indicating that it can effectively learn evaluation rules from a few examples. However, beyond three shots, additional examples offer limited gains and may even degrade accuracy. This may be attributed to prompt overload: too many shots increase input complexity and dilute the relevance of each shot, thereby confusing LLMs. Additionally, we observe that the scale of each shot significantly affects accuracy, as discussed in Appendix~\ref{appendix_shot_scale}, while the effect of shot order is negligible.

\textbf{Impact of Model Types.}
The models evaluated above are all general-purpose models. To explore the performance of reasoning-oriented models, we include GPT-5 with the most effective prompt for comparison. Fig.~\ref{radar}(b) shows that GPT-5 performs better than general-purpose models, likely because it excels at decomposing complex tasks into intermediate logical steps. This is particularly beneficial for tasks such as understanding composability, which requires trust aggregation across multiple trust chains. However, this improvement comes at the cost of substantially longer inference latency, limiting its practicality for large-scale trust evaluation.

\begin{tcolorbox}[
    colback=white,
    colframe=black,
    boxrule=1pt,
    arc=0pt,
    left=3pt,
    right=3pt,
    top=3pt,
    bottom=3pt,
    boxsep=0pt
]
\textbf{Finding 2.} LLMs' ability to understand trust properties is mainly affected by graph size and temporal complexity, while being largely insensitive to graph generators, trust level range, and graph density. Moderate shot numbers and reasoning-oriented models further improve accuracy.
\end{tcolorbox}

\section{Evaluation on LLMs' Capability for Trust Evaluation (RQ2)} \label{evaluation_rq2}
In this section, we investigate whether LLMs can perform trust evaluation on real-world datasets and compare their performance with state-of-the-art trust evaluation methods. We further analyze the explainability and robustness of LLM-based trust evaluation.

\subsection{Evaluation Setup}
\textbf{Evaluation Metrics.}
For static datasets, following reference~\cite{lin2020guardian}, we adopt F1-micro and Mean Absolute Error (MAE) to evaluate the performance of static trust evaluation methods. Compared with F1-micro, MAE is more suitable in this setting, as it reflects the numerical distance between predicted and ground-truth trust levels, rather than simply treating predictions as correct or incorrect. For dynamic datasets, which exhibit highly imbalanced trust distributions, we follow reference~\cite{wang2024trustguard} and employ F1-macro and Balanced Accuracy (BA) to ensure fair evaluation across the two trust levels. Note that higher values of F1-micro, F1-macro, and BA, while lower values of MAE, indicate better performance. The definitions of the four metrics are introduced in Appendix~\ref{evaluation_metrics}.

\textbf{Baselines.}
For static datasets, we select seven static trust evaluation methods as baselines: one statistics-based method (MoleTrust~\cite{massa2005controversial}), one inference-based method (OpinionWalk~\cite{liu2017opinionwalk}), and five ML-based methods (Matri~\cite{yao2013matri}, NeuralWalk~\cite{liu2019neuralwalk}, Guardian~\cite{lin2020guardian}, GATrust~\cite{jiang2022gatrust}, and TrustGNN~\cite{huo2024trustgnn}).  
For dynamic datasets, we focus on three state-of-the-art dynamic trust evaluation methods: Medley~\cite{lin2021medley}, DTrust~\cite{wen2023dtrust}, and TrustGuard~\cite{wang2024trustguard}. All of them are built on GNNs and consider the temporal dynamics of trust, demonstrating clear advantages over static methods. 
Overall, these baselines encompass various types of trust evaluation methods and cover both static and dynamic settings, enabling a comprehensive comparison. Their detailed descriptions are provided in Appendix~\ref{appendix_baseline}.

\textbf{Implementation Details.}
Based on the findings of \textbf{RQ1}, we select the three strongest general-purpose LLMs (Claude-3.7-Sonnet, Llama-4-Maverick, and DeepSeek-V3) with the best-performing prompt ``few-shot+knowledge+role'' for practical trust evaluation, as they offer high accuracy with acceptable inference time. Claude-3.7-Sonnet serves as the default model for parameter studies due to its superior understanding of trust properties. For static datasets, we use an 80\%/20\% train-test split and reserve 5\% of the training portion as a held-out validation set for input format construction. For dynamic datasets, we sort the data chronologically before applying the same split. The validation set is excluded from model training and few-shot demonstration selection. Following~\cite{liu2024exploring}, we randomly select three demonstration examples from the training set for each test sample to construct the ``few-shot'' prompt. Due to the high cost of querying LLM APIs, evaluating the entire validation and test sets is infeasible~\cite{liu2024exploring,chen2024exploring}. Instead, we first randomly sample 20 trustor-trustee pairs from the validation set for small-scale evaluation to determine the best input format, including the hop range and subgraph size. Then, we perform large-scale evaluation on 1,000 randomly sampled pairs (the same sample size used in existing work~\cite{lin2020guardian}). To ensure statistical significance, each evaluation is repeated five times with different random seeds; detailed statistical analysis is provided in Appendix~\ref{appendix_statistical}. Baseline methods are implemented in PyTorch following their original settings. All experiments are conducted on a server equipped with an Intel Xeon Platinum 8352Y CPU and two RTX 2080 Ti GPUs.

\subsection{Static Trust Graphs}

\subsubsection{\textbf{Input Format Study}} \label{input_format_study}
Given the limited context windows of LLMs, it is crucial to determine which parts of the trust graph should be included in a prompt. When evaluating the trust relationship of a trustor-trustee pair, not all graph information is relevant. As such, we construct an input subgraph by retaining nodes and edges along paths connecting the trustor and the trustee. This raises two design questions: (i) how many hops to consider, and (ii) how large the resulting subgraph should be. We conduct experiments to explore the impact of both parameters and validate the rationality of our input format design through an ablation study in Appendix~\ref{ablation_study}.

As shown in Fig.~\ref{input_static}, using fewer hops leads to poor performance due to the loss of critical information, whereas too many hops may introduce noise that degrades accuracy. Regarding subgraph size, the optimal settings vary significantly across the three datasets, suggesting that they reflect distinct trust scenarios. Overall, we use 3-hop subgraphs between the trustor and the trustee for Advogato and PGP, limiting their sizes to 60 and 5 edges, respectively. For Epinions, a 4-hop subgraph limited to 10 edges yields the best performance. These configurations offer a practical trade-off between information richness and input constraints.

\begin{figure}[tb]
    \centering
    \includegraphics[width=0.5\textwidth]{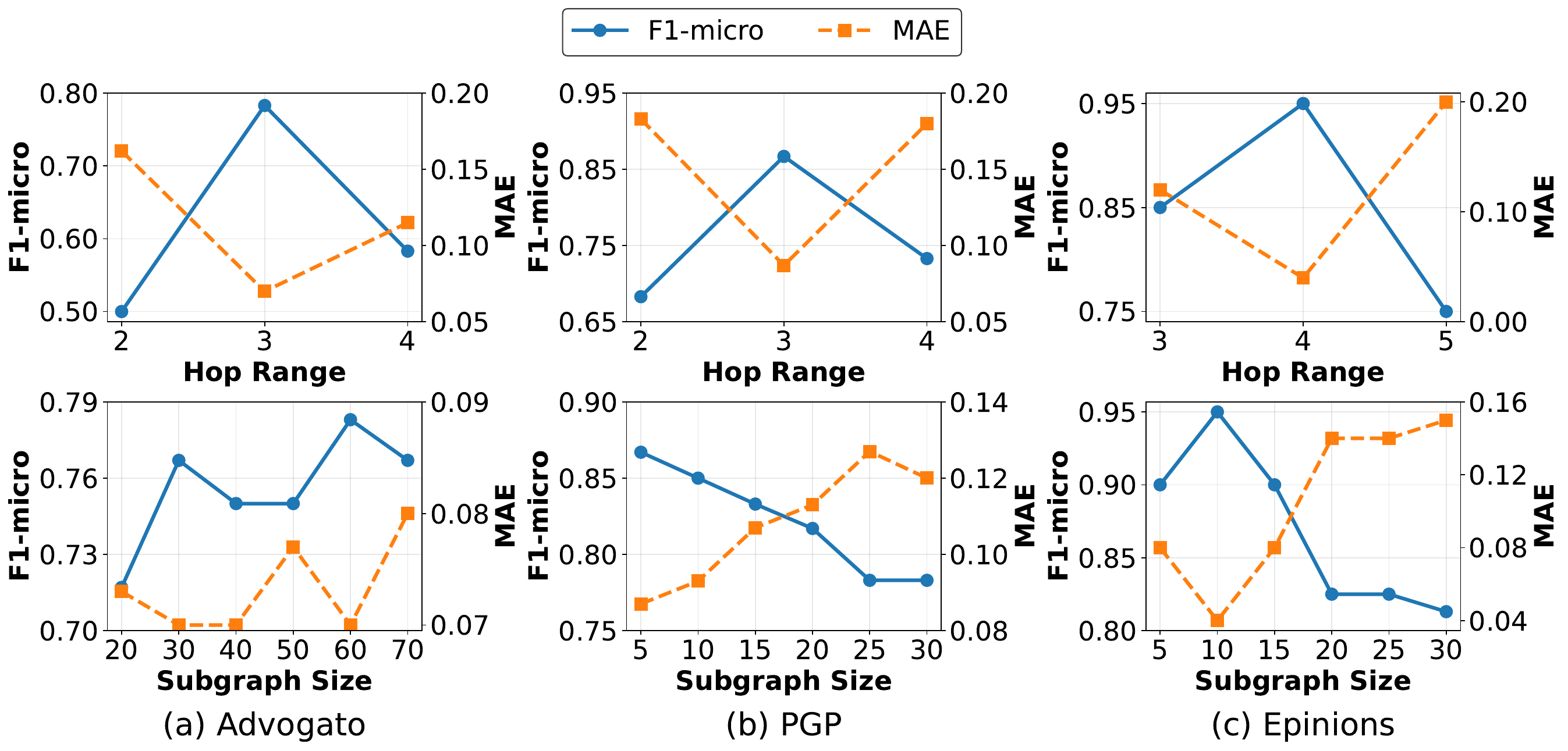}
    \vspace{-5mm}
    \caption{Impact of hop range and subgraph size on static datasets.}
    \label{input_static}
    \vspace{-3mm}
\end{figure}

\subsubsection{\textbf{Main Results}}
We compare three LLMs with seven representative trust evaluation methods on three static datasets. As shown in Table~\ref{llm_vs_baselines}, under the same 3-shot setting, LLMs consistently outperform 3-shot GNNs by substantial margins. The best-performing LLMs achieve F1-micro scores of 0.673, 0.850, and 0.881 on Advogato, PGP, and Epinions, respectively, compared with 0.518, 0.714, and 0.783 for the best 3-shot GNNs. The LLMs also outperform the non-learning-based methods and Matri across all three datasets. These results suggest that LLMs can learn effective trust evaluation rules from minimal supervision, whereas baselines relying on hand-crafted rules or simple architectures struggle to adapt to the complexity of real-world trust graphs.

\begin{table}[tbp]
  \centering
  \caption{Performance comparison between LLMs and baselines on static datasets.}
  \label{llm_vs_baselines}
  \renewcommand{\arraystretch}{1.1}
  \resizebox{\linewidth}{!}{
  \begin{threeparttable}
    \begin{tabular}{llcc|cc|cc}
      \toprule
      \multirow{2.5}{*}{\textbf{Category}}
      & \multirow{2.5}{*}{\textbf{Method}}
      & \multicolumn{2}{c|}{\textbf{Advogato}}
      & \multicolumn{2}{c|}{\textbf{PGP}}
      & \multicolumn{2}{c}{\textbf{Epinions}} \\
      \cmidrule(lr){3-4}
      \cmidrule(lr){5-6}
      \cmidrule(lr){7-8}
      &
      & \textbf{F1-micro$\uparrow$} & \textbf{MAE$\downarrow$}
      & \textbf{F1-micro$\uparrow$} & \textbf{MAE$\downarrow$}
      & \textbf{F1-micro$\uparrow$} & \textbf{MAE$\downarrow$} \\
      \midrule

      \multirow{2}{*}{\makecell[l]{Non-\\learning}}
      & MoleTrust~\cite{massa2005controversial}
      & 0.584 & 0.309 & 0.640 & 0.332 & 0.691 & 0.364 \\
      & OpinionWalk~\cite{liu2017opinionwalk}
      & 0.633 & 0.232 & 0.668 & 0.251 & 0.761 & 0.269 \\

      \midrule

      \multirow{5}{*}{\makecell[l]{Fully\\Supervised}}
      & Matri~\cite{yao2013matri}
      & 0.650 & 0.141 & 0.673 & 0.136 & 0.776 & 0.150 \\
      & NeuralWalk~\cite{liu2019neuralwalk}
      & \underline{0.740} & \underline{0.082}
      & \multicolumn{2}{c|}{Out of memory}
      & \multicolumn{2}{c}{Out of memory} \\
      & Guardian~\cite{lin2020guardian}
      & 0.730 & 0.087 & 0.870 & \underline{0.084} & 0.879 & 0.097 \\
      & GATrust~\cite{jiang2022gatrust}
      & 0.732 & 0.087 & \underline{0.871} & \underline{0.084} & \underline{0.880} & \underline{0.096} \\
      & TrustGNN~\cite{huo2024trustgnn}
      & \textbf{0.744} & \textbf{0.081}
      & \textbf{0.872} & \textbf{0.083}
      & \textbf{0.881} & \textbf{0.095} \\

      \midrule

      \multirow{3}{*}{\makecell[l]{3-shot\\GNNs}}
      & Guardian~\cite{lin2020guardian}
      & 0.446 & 0.202 & 0.651 & 0.258 & 0.783 & 0.173 \\
      & GATrust~\cite{jiang2022gatrust}
      & 0.518 & 0.159 & 0.714 & 0.203 & 0.622 & 0.302 \\
      & TrustGNN~\cite{huo2024trustgnn}
      & 0.398 & 0.196 & 0.546 & 0.332 & 0.556 & 0.355 \\

      \midrule

      \multirow{3}{*}{\makecell[l]{3-shot\\LLMs}}
      & DeepSeek-V3
      & 0.669 & 0.101 & 0.850 & 0.097
      & \textbf{0.881} & \textbf{0.095} \\
      & Llama-4-Maverick
      & 0.673 & 0.100 & 0.846 & 0.100
      & 0.854 & 0.117 \\
      & Claude-3.7-Sonnet
      & 0.672 & 0.101 & 0.849 & 0.096
      & \underline{0.880} & \underline{0.096} \\

      \bottomrule
    \end{tabular}

    \begin{tablenotes}[para,flushleft]
      \item[] The best and second-best results are in \textbf{bold} and
      \underline{underlined}, respectively.\\
      \item[] \textbf{Note:} Fully supervised learning-based methods use 80\% of the labeled data for training, whereas 3-shot GNNs and 3-shot LLMs use only three labeled examples.
    \end{tablenotes}
  \end{threeparttable}
  }
  \vspace{-3mm}
\end{table}

Compared with fully supervised learning-based methods, particularly GNN-based approaches, LLMs achieve state-of-the-art performance on Epinions while remaining less accurate on Advogato and PGP. This indicates that learning-based methods can extract accurate trust patterns when sufficient labeled data are available, with GNNs further benefiting from their ability to handle complex structural relationships, which are critical for evaluating trust between node pairs. However, in practice, trust labels are difficult to obtain and often require subjective ratings from thousands of users, devices, or other entities~\cite{hou2022handling,wang2020survey}, which restricts the practicality of fully supervised methods.

\begin{figure}[tb]
    \centering
    \includegraphics[width=0.48\textwidth]{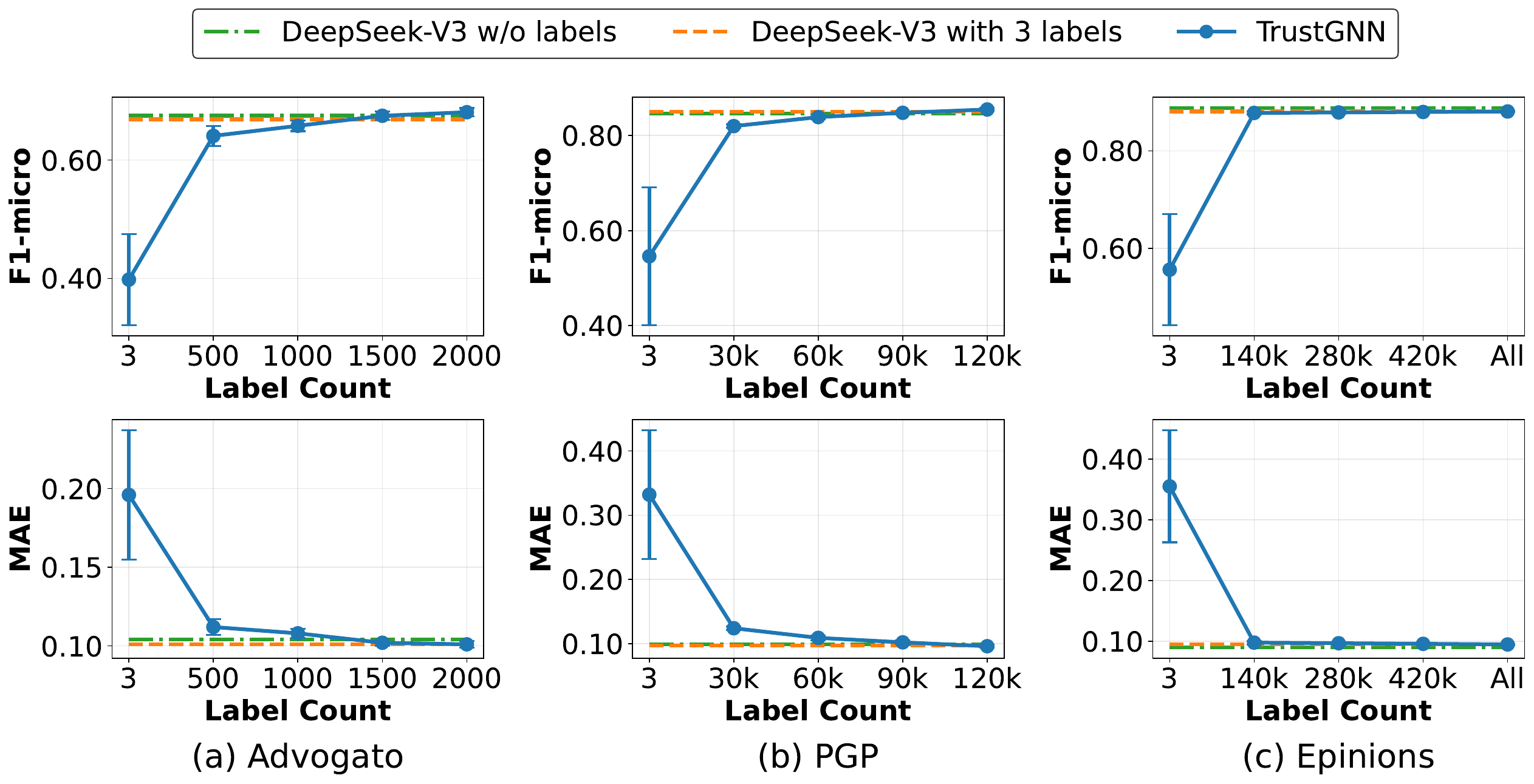}
    \caption{Performance comparison between DeepSeek-V3 and TrustGNN under limited supervision on static datasets.}
    \label{label_static}
    \vspace{-4mm}
\end{figure}

To further investigate the impact of label availability, we compare DeepSeek-V3 and TrustGNN (the strongest LLM and non-LLM methods on static datasets) with varying numbers of labels. Fig.~\ref{label_static} shows that DeepSeek-V3 has a clear advantage over TrustGNN under extremely limited supervision, achieving performance comparable to TrustGNN trained with about 1,500 labels on Advogato, 90,000 labels on PGP, and even all available labels (569,206) on Epinions. As the number of labels increases, TrustGNN's accuracy improves while its standard deviation decreases. This indicates that learning-based methods are highly dependent on label availability. Surprisingly, DeepSeek-V3 without labels performs slightly better than its few-shot counterpart, serving as a strong baseline. We attribute this to two factors: (i) knowledge acquired during pretraining generalizes well to trust evaluation, and (ii) few-shot prompting can introduce variance when the selected demonstration examples are unrepresentative. Overall, these results highlight LLMs' strong zero-shot and few-shot capabilities, as well as their promise for real-world applications.

\begin{tcolorbox}[
    colback=white,
    colframe=black,
    boxrule=1pt,
    arc=0pt,
    left=3pt,
    right=3pt,
    top=3pt,
    bottom=3pt,
    boxsep=0pt
]
\textbf{Finding 3.} On static datasets, LLMs consistently outperform non-learning-based approaches but remain less accurate than learning-based methods (especially GNNs) when sufficient labels are available. Under limited supervision, however, LLMs show a clear advantage, exhibiting strong generalization and practicality.
\end{tcolorbox}

\subsubsection{\textbf{Explainability Analysis}} \label{explainability_analysis}
\begin{figure}[tb]
    \centering
    \includegraphics[width=0.26\textwidth]{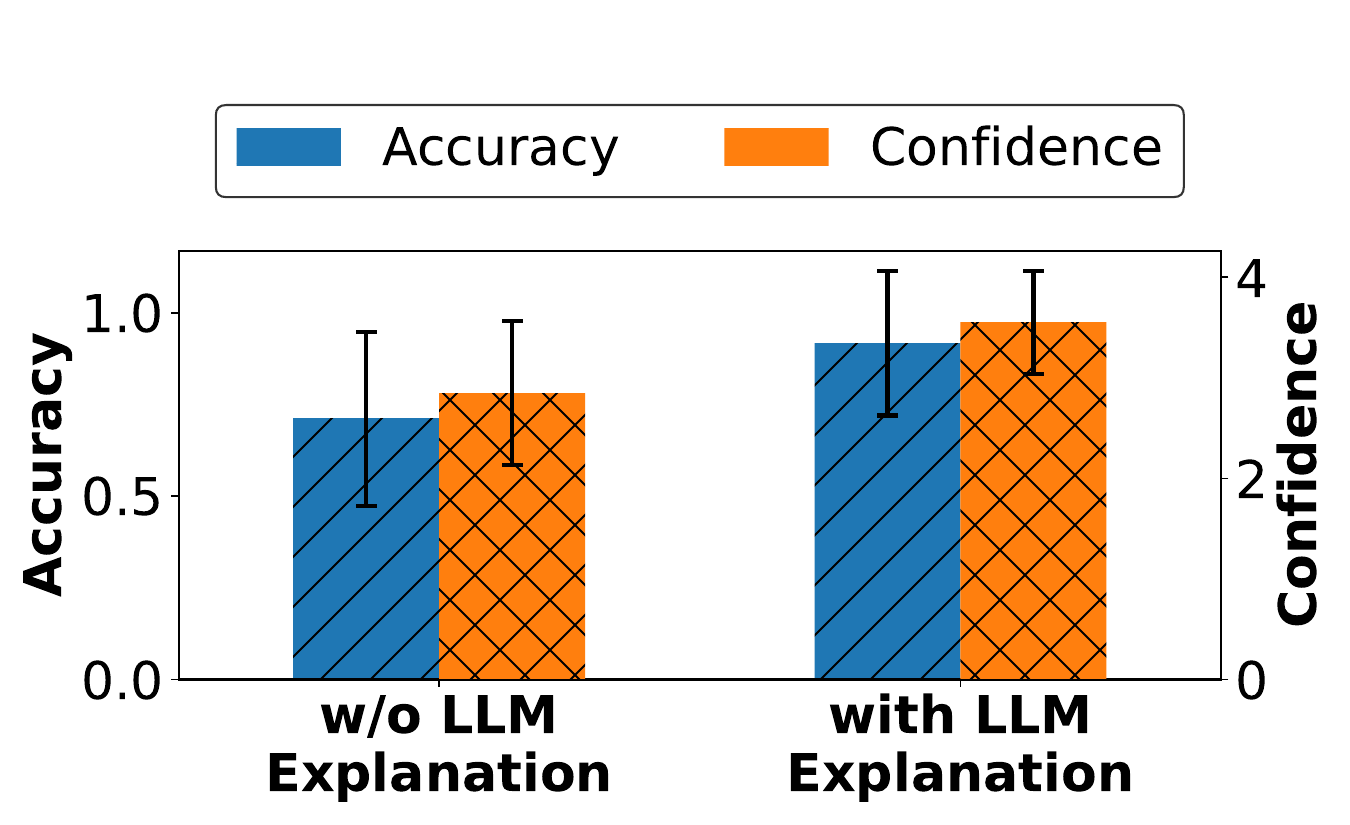}
    \vspace{-1mm}
    \caption{Results of user studies on LLM explanations.}
    \label{user_study}
    \vspace{-4mm}
\end{figure}

Using LLMs for trust evaluation offers a distinct advantage over traditional ML methods: LLMs can explicitly articulate their inference processes step by step, providing clear justifications for trust evaluation results, as illustrated in Fig.~\ref{case_static}. This level of explainability is particularly valuable in practice, as it can enhance user trust and acceptance~\cite{shin2021effects} and help refine trust evaluation rules. 
To investigate the usefulness of these explanations, we conducted a user study (see Appendix~\ref{appendix_user_study} for details) involving 20 volunteer participants. The participants were asked to answer 12 trust evaluation questions under two conditions: with and without LLM-generated explanations. They also reported their confidence on each answer using a 4-point scale. The results in Fig.~\ref{user_study} show that providing explanations improves the participants' trust evaluation accuracy by 28.97\% and increases their reported confidence by 24.73\% compared with the no-explanation condition. Consequently, LLM-generated explanations not only make trust evaluation transparent, but also help users make accurate and confident trust decisions.

\begin{tcolorbox}[
    colback=white,
    colframe=black,
    boxrule=1pt,
    arc=0pt,
    left=3pt,
    right=3pt,
    top=3pt,
    bottom=3pt,
    boxsep=0pt
]
\textbf{Finding 4.} LLMs' step-by-step reasoning significantly improves both decision quality and human understanding of trust evaluation results.
\end{tcolorbox}

\subsection{Dynamic Trust Graphs}

\subsubsection{\textbf{Input Format Study}}
Using the same methodology as in Section~\ref{input_format_study}, we find that Bitcoin-OTC and Bitcoin-Alpha share the same optimal parameter settings: a 3-hop subgraph limited to 50 edges, likely due to their structural similarity. We also validate the effectiveness of our input format design for dynamic trust graphs, including both structural and temporal principles. Detailed results are provided in Appendix~\ref{ablation_study}.

\begin{table}[tb]
  \centering
  \footnotesize
  \caption{Performance comparison between LLMs and baselines on dynamic datasets.}
  \label{llm_vs_baselines_dynamic}
  \renewcommand{\arraystretch}{1.1}
\resizebox{\linewidth}{!}{
\begin{threeparttable}
  \begin{tabular}{llcc|cc}
    \toprule
    \multirow{2.5}{*}{\textbf{Category}}
    & \multirow{2.5}{*}{\textbf{Method}}
    & \multicolumn{2}{c|}{\textbf{Bitcoin-OTC}}
    & \multicolumn{2}{c}{\textbf{Bitcoin-Alpha}} \\
    \cmidrule(lr){3-4} \cmidrule(lr){5-6}
    &
    & \textbf{F1-macro$\uparrow$}
    & \textbf{BA$\uparrow$}
    & \textbf{F1-macro$\uparrow$}
    & \textbf{BA$\uparrow$} \\
    \midrule

    \multirow{6}{*}{\makecell[l]{Fully\\Supervised}}
    & Guardian~\cite{lin2020guardian}
    & 0.613 & 0.613 & 0.530 & 0.540 \\
    & GATrust~\cite{jiang2022gatrust}
    & 0.581 & 0.574 & 0.512 & 0.521 \\
    & TrustGNN~\cite{huo2024trustgnn}
    & 0.624 & 0.627 & 0.538 & 0.535 \\
    & Medley~\cite{lin2021medley}
    & 0.622 & 0.595 & 0.571 & 0.558 \\
    & DTrust~\cite{wen2023dtrust}
    & 0.637 & \underline{0.658} & 0.583 & \textbf{0.607} \\
    & TrustGuard~\cite{wang2024trustguard}
    & \textbf{0.682} & \textbf{0.676} & \underline{0.590} & 0.591 \\

    \midrule

    \multirow{3}{*}{\shortstack[l]{3-shot\\GNNs}}
    & Medley~\cite{lin2021medley}
    & 0.496 & 0.512 & 0.493 & 0.507 \\
    & DTrust~\cite{wen2023dtrust}
    & 0.436 & 0.488 & 0.477 & 0.502 \\
    & TrustGuard~\cite{wang2024trustguard}
    & 0.474 & 0.497 & 0.493 & 0.527 \\

    \midrule

    \multirow{3}{*}{\shortstack[l]{3-shot\\LLMs}}
    & DeepSeek-V3
    & \underline{0.678} & 0.652 & \textbf{0.609} & \underline{0.594} \\
    & Llama-4-Maverick
    & 0.524 & 0.525 & 0.571 & 0.562 \\
    & Claude-3.7-Sonnet
    & 0.619 & 0.643 & 0.532 & 0.545 \\

    \bottomrule
  \end{tabular}

\begin{tablenotes}[para,flushleft]
\item[] \textbf{Note:} Fully supervised learning-based methods use 80\% of the labeled data for training, whereas 3-shot GNNs and 3-shot LLMs use only three labeled examples.
\end{tablenotes}

\end{threeparttable}
}
  \vspace{-4mm}
\end{table}

\subsubsection{\textbf{Main Results}}
We compare LLMs with three top-performing static trust evaluation methods (Guardian, GATrust, and TrustGNN) and three state-of-the-art dynamic methods (Medley, DTrust, and TrustGuard). As shown in Table~\ref{llm_vs_baselines_dynamic}, all methods struggle to achieve high performance on two dynamic datasets, particularly on the highly imbalanced Bitcoin-Alpha dataset. This highlights the challenges of real-world trust evaluation, where ``distrust'' labels are scarce and trust relationships evolve over time. Due to these difficulties, we observe substantial performance variation across LLMs. Nevertheless, under the same 3-shot setting, all three LLMs consistently outperform the dynamic trust evaluation methods, indicating their strong few-shot capability. Compared with fully supervised methods, most LLMs also perform better than static methods, likely because they can incorporate temporal information into trust reasoning. Among them, DeepSeek-V3 performs best, achieving performance comparable to TrustGuard while outperforming the other baselines. These results highlight the potential of LLMs for trust evaluation in practice.

\begin{figure}[tb]
    \centering
    \includegraphics[width=0.43\textwidth]{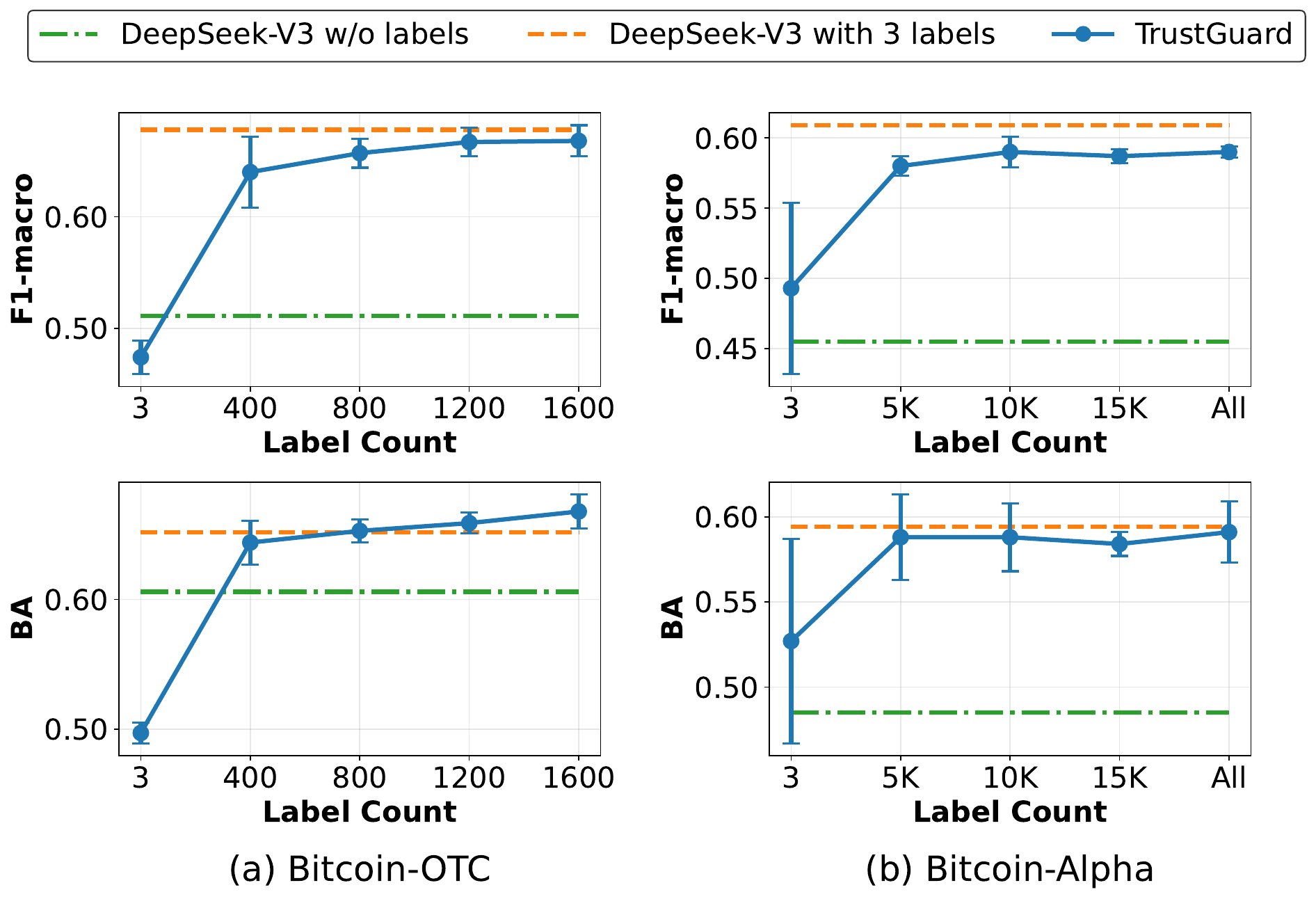}
    \vspace{-0.5mm}
    \caption{Performance comparison between DeepSeek-V3 and TrustGuard under limited supervision on dynamic datasets.}
    \label{label_dynamic}
    \vspace{-2mm}
\end{figure}

Under limited supervision, Fig.~\ref{label_dynamic} shows that TrustGuard becomes highly unstable, likely because its attention-based architecture requires substantial labeled data for effective training. Moreover, DeepSeek-V3 without labels is not competitive on either dataset due to the complexity of dynamic trust evaluation. In contrast, with only three labeled examples, DeepSeek-V3 matches the performance of TrustGuard trained with about 1,200 labels on Bitcoin-OTC and with all available labels (19,348) on Bitcoin-Alpha. These results further demonstrate LLMs' superiority under scarce supervision.

\begin{tcolorbox}[
    colback=white,
    colframe=black,
    boxrule=1pt,
    arc=0pt,
    left=3pt,
    right=3pt,
    top=3pt,
    bottom=3pt,
    boxsep=0pt
]
\textbf{Finding 5.} Dynamic and imbalanced trust graphs pose challenges for trust evaluation, yet DeepSeek-V3 achieves competitive performance with state-of-the-art fully supervised methods using only three labeled examples.
\end{tcolorbox}

\subsubsection{\textbf{Robustness Analysis}}
We consider that attackers can manipulate LLM inputs, including trust graphs in the user input and demonstration examples used for few-shot prompting, with the goal of inducing incorrect trust evaluation results. However, they cannot access or modify the LLM architecture or parameters, reflecting practical API-based deployments. These assumptions are realistic, as trust graphs may be poisoned by malicious interactions, while few-shot demonstrations may be selected from compromised data sources. General LLM security attacks (e.g., prompt injection) are beyond the scope of this work.

Based on this threat model, we consider attacks targeting trust graphs and few-shot demonstrations. For the former, we study bad/good-mouthing attacks~\cite{wang2024trustguard,wang2025cat} and their temporal variants. For each trustor-trustee pair to be evaluated, we randomly flip a portion of trust relationships (e.g., from trust to distrust) within the associated subgraph, termed \textit{mix attacks}. For the temporal variants, we adjust the timestamps of malicious trust relationships to match the prediction time, exploiting the fact that recent interactions receive significant attention in dynamic trust evaluation~\cite{lin2021medley,wang2025cat}. We refer to these attacks as \textit{time-aware attacks}. For attacks on few-shot demonstrations, we modify the ground truth provided in demonstration examples to mislead few-shot learning, referred to as \textit{shot attacks}. We evaluate each attack under both low- and high-perturbation settings and report only F1-macro due to space constraints. For instance, 10\%/30\% perturbation means modifying the corresponding fraction of edges in the subgraph. More results are provided in Appendix~\ref{appendix_robustness}.


\begin{figure}[tb]
    \centering
    \includegraphics[width=0.48\textwidth]{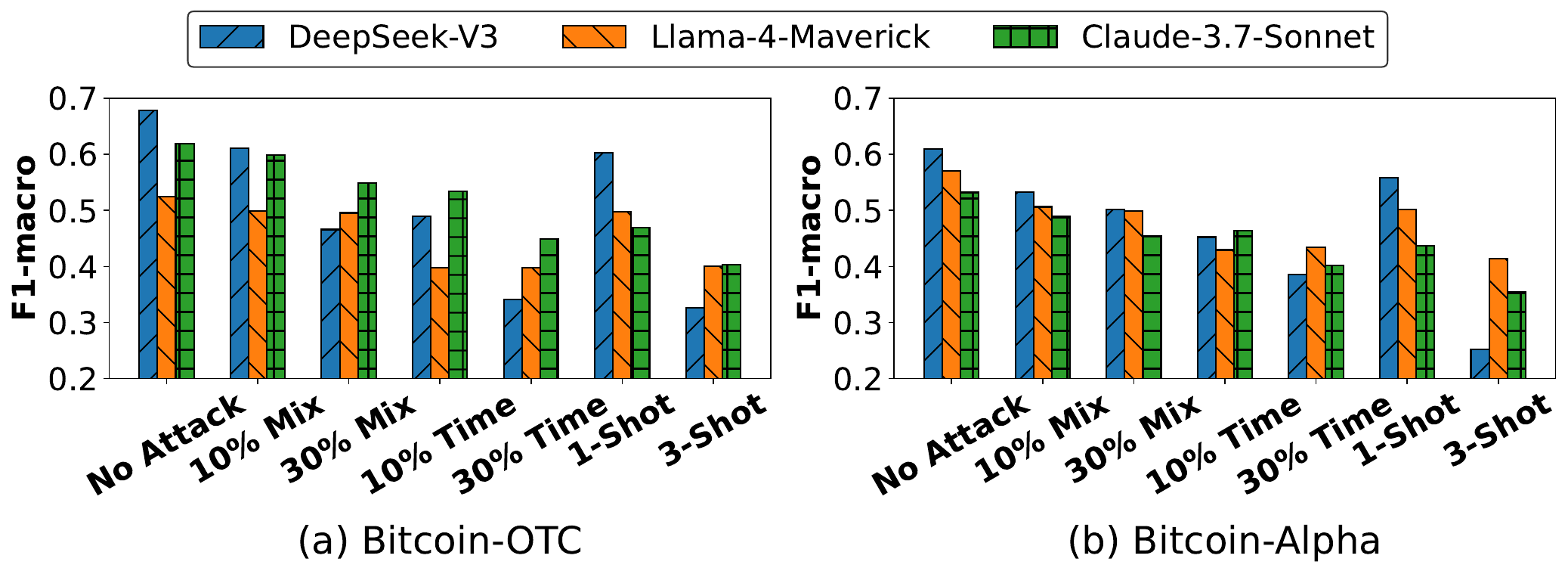}
    \vspace{-2mm}
    \caption{Robustness comparison against mix, time-aware, and shot attacks under different perturbation levels.}
    \label{robustness_analysis}
    \vspace{-3mm}
\end{figure}

As shown in Fig.~\ref{robustness_analysis}, all LLMs suffer significant performance drops under the three attacks, especially at high perturbation levels. For instance, on Bitcoin-Alpha, DeepSeek-V3 experiences absolute F1-macro drops of 10.79\%, 22.30\%, and 35.66\% under 30\% mix attacks, 30\% time-aware attacks, and 3-shot attacks, respectively. Among these, the latter two attacks are the most damaging. To understand the underlying causes, we analyze the inference behavior of LLMs. We find that time-aware attacks indeed change the prediction patterns of LLMs by forcing them to focus on recent interactions. For instance, under such attacks, one prediction by Llama-4-Maverick states that \textit{``The presence of multiple edges with trust level 0 at the given time suggests that the trust level might be leaning towards 0 for new or unobserved nodes.''} Regarding 3-shot attacks, the false demonstration examples make LLMs employ a conservative prediction strategy that tends to output the ``distrust'' answer, as illustrated in Fig.~\ref{robustness_ratio}. These findings motivate improving the quality of prompts, including demonstrations and trust evaluation queries.

We propose a lightweight input-level defense that integrates adversarial example augmentation, temporal edge filtering, and consistency correction (see Appendix~\ref{appendix_robustness} for their descriptions and design justifications). Table~\ref{defense_gain} shows that the combined defense consistently improves the absolute performance under all three types of attacks and is particularly effective against time-aware attacks, recovering performance by over 13\% on average. Furthermore, this defense is easy to deploy and broadly applicable to any LLM-based trust evaluation.

\begin{table}[tb]
  \centering
  \footnotesize
  \caption{Absolute performance improvements achieved by the defense under high-perturbation attacks.}
  \label{defense_gain}
\renewcommand{\arraystretch}{1.1}
\resizebox{\linewidth}{!}{
  \begin{tabular}{lccc|ccc}
    \toprule
    \multirow{2.5}{*}{\textbf{Method}} &
    \multicolumn{3}{c|}{\textbf{Bitcoin-OTC}} &
    \multicolumn{3}{c}{\textbf{Bitcoin-Alpha}} \\
    \cmidrule(lr){2-4} \cmidrule(lr){5-7}
    & \textbf{30\% Mix} & \textbf{30\% Time} & \textbf{3-Shot}
    & \textbf{30\% Mix} & \textbf{30\% Time} & \textbf{3-Shot} \\
    \midrule
    DeepSeek-V3       & 5.53\% & 23.18\%  & 8.01\% & 3.27\% & 14.72\% & 29.08\% \\
    Llama-4-Maverick  & 2.55\%  & 12.19\%  & 1.76\%  & 1.54\% & 9.65\% & 10.91\% \\
    Claude-3.7-Sonnet & 2.72\%  & 14.04\%  & 1.81\%  & 3.91\% & 9.50\% & 11.02\% \\
    \bottomrule
  \end{tabular}
}
  \vspace{-3mm}
\end{table}

\begin{tcolorbox}[
    colback=white,
    colframe=black,
    boxrule=1pt,
    arc=0pt,
    left=3pt,
    right=3pt,
    top=3pt,
    bottom=3pt,
    boxsep=0pt
]
\textbf{Finding 6.} LLMs are vulnerable to attacks targeting trust evaluation and demonstration examples, while the proposed input-level defense effectively mitigates their impact.
\end{tcolorbox}

\section{Discussion and Limitations} \label{discussion}
\textbf{Cost and Efficiency Analysis.} The total API cost for obtaining the main results reported in Tables~\ref{main_results_rq1},~\ref{llm_vs_baselines}, and~\ref{llm_vs_baselines_dynamic} is approximately \$983.68, as detailed in Appendix~\ref{appendix_cost}. The appendix further shows that the efficiency of LLM-based trust evaluation is primarily influenced by inference time. Specifically, LLMs typically require several seconds to evaluate a single trustor-trustee pair, whereas learning-based approaches can evaluate thousands of pairs in less than one second. To mitigate this efficiency gap, we adopt batch processing by grouping multiple trustor-trustee pairs into a single query. Table~\ref{inference_time} reports the inference time and F1-micro of DeepSeek-V3 on two large datasets under different batch sizes. The results show a clear efficiency-accuracy trade-off: large batch sizes substantially reduce inference time per evaluation pair, but may slightly degrade accuracy. This is likely because longer prompts introduce more irrelevant content and reduce the effective attention that LLMs pay to each evaluation pair. Moreover, LLMs' step-by-step reasoning often disappears at large batch sizes, indicating that the high inference cost at $batch\ size=1$ mainly comes from generating explanations for each trust evaluation result.

\begin{tcolorbox}[
    colback=white,
    colframe=black,
    boxrule=1pt,
    arc=0pt,
    left=3pt,
    right=3pt,
    top=3pt,
    bottom=3pt,
    boxsep=0pt
]
\textbf{Finding 7.} LLM efficiency is mainly determined by inference time, and batch processing improves efficiency at the cost of some accuracy and explainability.
\end{tcolorbox}

\textbf{Real-World Use Case.} Previous experiments show that LLMs are particularly effective in label-scarce trust evaluation scenarios, making them suitable for newly deployed systems with limited labeled data. A concrete example is trust-based fraud detection in early Bitcoin trading platforms. Bitcoin-Alpha and Bitcoin-OTC~\cite{kumar2016edge,kumar2018rev2} contain 3,783/5,881 users and 24,186/35,592 timestamped trust ratings, but only 54/77 ratings are available in the first 30 days and 195/292 in the first 90 days. This creates a three-month label-scarce period in which supervised GNNs are difficult to train reliably, as shown in Fig.~\ref{label_dynamic}. During this stage, DeepSeek-V3 can serve as an initial trust evaluator that requires little task-specific supervision. For about 1,000 queried trustor-trustee pairs, it achieves strong performance at an API cost of only \$0.40/\$0.45 on Bitcoin-Alpha/OTC, with an average inference time of 10.24/5.66 seconds per query, while also providing evaluation justifications. As labeled data accumulate, learning-based approaches, such as TrustGuard~\cite{wang2024trustguard}, become more attractive because they can fully exploit abundant supervision to achieve high accuracy while maintaining efficiency. LLM outputs can also serve as pseudo-labels to bootstrap ML models. Recent advances in agent-based technologies further enable hybrid LLM-GNN systems that dynamically select or combine trust evaluators based on data availability~\cite{du2025graphmaster}.

\begin{table}[t]
\centering
\footnotesize
\caption{Impact of batch size on inference time and F1.}
\label{inference_time}

\begin{tabular}{llccccc}
\toprule
\multirow{2}{*}{\textbf{Dataset}} & \multirow{2}{*}{\textbf{Metric}}
& \multicolumn{5}{c}{\textbf{Batch size}} \\
\cmidrule(lr){3-7}
 &  & \textbf{1} & \textbf{5} & \textbf{10} & \textbf{20} & \textbf{40} \\
\midrule
\multirow{2}{*}{PGP}
 & F1-micro        & 0.850 & 0.794 & 0.799 & 0.778 & 0.775 \\
 & Time (s)        & 4.821 & 2.825 & 3.860 & 6.940 & 11.960 \\
\midrule
\multirow{2}{*}{Epinions}
 & F1-micro        & 0.881 & 0.879 & 0.871 & 0.874 & 0.851 \\
 & Time (s)        & 9.984 & 2.855 & 4.160 & 6.480 & 11.720 \\
\bottomrule
\end{tabular}
\vspace{-3mm}
\end{table}

\textbf{Limitations.} LLM4Trust has two limitations: \textit{Bias in Expert Knowledge} and \textit{Lack of Evaluation on Text-Attributed Trust Graphs}. First, trust is a complex concept that varies across different scenarios and even among individuals. Therefore, it is quite difficult to define general trust evaluation rules suitable for all scenarios. That is, some of the expert knowledge used to guide LLM reasoning may be biased in certain situations. This can also explain why Claude-3.7-Sonnet achieves the best performance in understanding basic trust properties, while DeepSeek-V3 performs better on real-world trust evaluation. 

Second, our experiments focus on graph-structured data (both synthetic and real-world), where trust is evaluated solely based on node interactions. However, trust is a multifaceted concept influenced by various factors. For instance, in online social networks, users' inherent attributes (e.g., hobbies and occupations) can also affect trust, as users with similar interests tend to trust each other~\cite{wang2021c}. Therefore, it is important to explore trust evaluation in Text-Attributed Graphs (TAGs), where nodes are associated with rich textual descriptions. We did not study such graphs due to the limited availability of high-quality textual data in existing datasets. Nevertheless, since LLMs excel at processing textual information, LLM4Trust can be naturally extended by incorporating node descriptions into prompts. We believe that such extensions would further improve the accuracy of LLM-based trust evaluation and leave this investigation as future work.



\section{Conclusion}
In this paper, we proposed LLM4Trust, the first benchmark framework that explores LLMs' capabilities for trust evaluation. We first evaluated eight LLMs with nine prompt methods on five basic trust properties and found that LLMs generally understand these properties well. We then applied the best LLM-prompt combinations to five real-world datasets, using two strategies to address LLMs' context window limitations. The results show that LLMs have strong potential for real-world trust evaluation, particularly under limited supervision, while their transparent inference processes enhance the explainability of trust evaluation results. However, LLMs remain vulnerable to attacks targeting trust graphs and demonstration examples, and incur high inference costs. Future work will develop stronger defenses and improve inference efficiency for practical deployment.


\section*{Ethical Considerations}
This work introduces LLM4Trust, a benchmark framework for assessing LLMs' trust evaluation capabilities. While designed to enhance cybersecurity, LLM4Trust may be misused to infer trust relationships for malicious purposes. We therefore recommend responsible use and appropriate safeguards, such as restricting access to authorized stakeholders.
The human-related studies in Section~\ref{explainability_analysis} were approved by the Institutional Review Board (IRB) of our institutes. All participants provided informed consent, and no unnecessary personal information was collected.

\section*{Acknowledgment}
This work is supported in part by the National Natural Science Foundation of China under Grants 62676303 and U23A20300; in part by the ``Pioneer'' and ``Leading Goose'' R\&D Program of Zhejiang under Grant No. 2026C01020; in part by the Fundamental Research Funds for the Central Universities under Grant QTZX26067; in part by the Xi'an Science and Technology Bureau and the Xi'an Science and Technology Project under Grant 25RKYJ0010, and in part by the 111 Center under Grant B16037.

\bibliographystyle{IEEEtran}
\bibliography{ref}

\appendices

\section{Algorithms for Generating Q\&A pairs} \label{appendix_algorithm}
Algorithms~\ref{algorithm_dynamicity}--\ref{algorithm_context} detail the procedures for generating question-answer pairs for the five fundamental trust properties.

\begin{algorithm}[!htbp]
\footnotesize
\SetCommentSty{small}
\LinesNumbered
\caption{Generating Q\&A pairs for dynamicity}
\label{algorithm_dynamicity}

\KwIn{Number of nodes $N$, edge creation probability $p$, time span $T$;}
\KwOut{Trust graph $\mathcal{G}$, query node pair, answer;}

\While{\textnormal{True}}{
    Generate a directed trust graph $\mathcal{G} = ER(N, p)$;

    Assign a timestamp $t \sim U(\{1, 2, \dots, T\})$ to each edge in $\mathcal{E}$;

    Randomly select a node pair $(u, v) \in \mathcal{V} \times \mathcal{V}$ such that $u \neq v$;

    \If{$\mathcal{G}.\textnormal{has\_path}(u, v)$}{
        $\mathcal{P} \leftarrow$ All paths from $u$ to $v$;

        $est\_times \leftarrow []$; \quad \textcolor[gray]{0.6}{// List of establishment times}

        \For{$path \in \mathcal{P}$}{
            $est\_time \leftarrow \max(\textnormal{timestamp}(e)), \forall e \in path$;

            Append $est\_time$ to $est\_times$;
        }

        $answer \leftarrow \min(est\_times)$;

        \Return{$\mathcal{G}$, $(u, v)$, $answer$}
    }
}
\end{algorithm}

\begin{algorithm}[!htbp]
\footnotesize
\SetCommentSty{small}
\LinesNumbered
\caption{Generating Q\&A pairs for asymmetry}
\label{algorithm_asymmetry}

\KwIn{Number of nodes $N$, edge creation probability $p$, number of trust levels $W$;}
\KwOut{Trust graph $\mathcal{G}$, query node pair, answer;}

\While{\textnormal{True}}{
    Generate a directed trust graph $\mathcal{G} = ER(N, p)$;

    Assign a trust level $w \sim U(\{1, 2, \dots, W\})$ to each edge in $\mathcal{E}$;

    Randomly select a node pair $(u, v) \in \mathcal{V} \times \mathcal{V}$ such that $u \neq v$;

    \If{$\mathcal{G}.\textnormal{has\_edge}(u, v)$ \textbf{and} $\neg\mathcal{G}.\textnormal{has\_edge}(v, u)$}{
        \If{$\neg\mathcal{G}.\textnormal{has\_path}(v, u)$}{
            \Return{$\mathcal{G}$, $(v, u)$, $0$}
        }
    }
}
\end{algorithm}

\begin{algorithm}[!htbp]
\footnotesize
\SetCommentSty{small}
\LinesNumbered
\caption{Generating Q\&A pairs for conditional transitivity}
\label{algorithm_transitivity}

\KwIn{Number of nodes $N$, edge creation probability $p$, number of trust levels $W$;}
\KwOut{Trust graph $\mathcal{G}$, query node pair, answer;}

\While{\textnormal{True}}{
    Generate a directed trust graph $\mathcal{G} = ER(N, p)$;

    Assign a trust level $w \sim U(\{1, 2, \dots, W\})$ to each edge in $\mathcal{E}$;

    Randomly select a node pair $(u, v) \in \mathcal{V} \times \mathcal{V}$ such that $u \neq v$;

    \If{$\mathcal{G}.\textnormal{has\_path}(u, v)$ \textbf{and} $\neg \mathcal{G}.\textnormal{has\_edge}(u, v)$}
    {
        $\mathcal{P} \leftarrow$ All paths from $u$ to $v$;
        
        \If{$|\mathcal{P}| = 1$}{
            $path \leftarrow$ the only element in $\mathcal{P}$;

            $min\_weight \leftarrow \min\{w(u,v) \mid (u,v) \in path\}$;

            \Return{$\mathcal{G}$, $(u, v)$, $min\_weight$}
        }

    }
}
\end{algorithm}

\begin{algorithm}[!htbp]
\footnotesize
\SetCommentSty{small}
\LinesNumbered
\caption{Generating Q\&A pairs for composability}
\label{algorithm_composability}

\KwIn{Number of nodes $N$, edge creation probability $p$, number of trust levels $W$;}
\KwOut{Trust graph $\mathcal{G}$, query node pair, answer;}

\While{\textnormal{True}}{
    Generate a directed trust graph $\mathcal{G} = ER(N, p)$;

    Assign a trust level $w \sim U(\{1, 2, \dots, W\})$ to each edge in $\mathcal{E}$;

    Randomly select a node pair $(u, v) \in \mathcal{V} \times \mathcal{V}$ such that $u \neq v$;

    \If{$\neg \mathcal{G}.\textnormal{has\_edge}(u, v)$}{
    $\mathcal{P} \leftarrow$ All paths from $u$ to $v$;

    \If{$|\mathcal{P}| > 1$}{
        $min\_weight, max\_weight \leftarrow \infty, -\infty$;

        \For{$path \in \mathcal{P}$}{
            $weight \leftarrow \min \{ w(u,v) \mid (u,v) \in path \}$;

            \If{$weight < min\_weight$}{
                $min\_weight \leftarrow weight$;
                
                $min\_path \leftarrow path$;
            }

            \If{$weight > max\_weight$}{
                $max\_weight \leftarrow weight$;
                
                $max\_path \leftarrow path$;
            }
        }

        \If{$min\_path \neq max\_path$}{
            \Return{$\mathcal{G}$, $(u, v)$, $[min\_weight, max\_weight]$}
        }
    }
}
}
\end{algorithm}

\begin{algorithm}[!htbp]
\footnotesize
\SetCommentSty{small}
\LinesNumbered
\caption{Generating Q\&A pairs for context-awareness}
\label{algorithm_context}

\KwIn{Number of nodes $N$, edge creation probability $p$, number of trust levels $W$, context set $C$;}
\KwOut{Trust graph $\mathcal{G}$, query node pair, answer;}

\While{\textnormal{True}}{
    Generate a directed trust graph $\mathcal{G} = ER(N, p)$;

    Assign trust levels $w \sim U(\{1, 2, \dots, W\})$ along with contexts $\{c_1,c_2,...,c_k\}$ to each edge in $\mathcal{E}$;

    Randomly select a node pair $(u, v) \in \mathcal{V} \times \mathcal{V}$ such that $u \neq v$;

    \If{$\mathcal{G}.\textnormal{has\_edge}(u, v)$}
    {
        $\mathcal{P} \leftarrow$ All paths from $u$ to $v$;
        
        \If{$|\mathcal{P}| = 1$}{

            \Return{$\mathcal{G}$, $(u, v, c_{new})$, $0$} \quad \textcolor[gray]{0.6}{// $c_{new}$ refers to a new context}
        }

    }
}
\end{algorithm}

\section{Domain Expert Knowledge} \label{appendix_knowledge}
Table~\ref{prompt_knowledge} presents domain expert knowledge used in the ``knowledge'' prompt.

\begin{table}[tbp]
\centering
\caption{Task-specific domain expert knowledge.}
\label{prompt_knowledge}
\renewcommand{\arraystretch}{1.1}
\begin{tabular}{>{\centering\arraybackslash}m{1.6cm} m{6.3cm}}
\toprule
\textbf{Task} & \textbf{Knowledge} \\
\midrule
Dynamicity & Trust is dynamic over time; specifically, $(u, v, t_1)$ is established earlier than $(u, v, t_2)$ if $t_1 < t_2$. \\
\midrule
Asymmetry & Trust is inherently asymmetric; specifically, the trust of node $u$ in node $v$, represented as $(u, v, w)$, does not imply an equivalent trust level of node $v$ in node $u$, which would be represented as $(v, u, w)$. \\
\midrule
Conditional Transitivity & Trust is propagative; specifically, the trust of node $u$ in node $v$ along a given path is determined by the minimum trust level among all edges constituting that path. \\
\midrule
Composability & Trust is composable; specifically, the trust of node $u$ in node $v$ along a single path is determined by the minimum trust level among all edges constituting that path. When there are multiple paths from node $u$ to node $v$, the trust level from $u$ to $v$ falls within the range defined by the individual path trust levels. \\
\midrule
Context-awareness & Trust is context-aware; specifically, $(u, v, c_1, w)$ means that node $u$ trusts node $v$ with a trust level of $w$ in the context of $c_1$, but it does not imply that $u$ has the same trust in $v$ in the context of $c_2$, which would be represented as $(u, v, c_2, w)$. \\
\bottomrule
\end{tabular}
\vspace{-1mm}
\end{table}

\section{Prompt Details} \label{appendix_prompt}
\begin{figure}[tbp]
    \centering
    \includegraphics[width=0.48\textwidth]{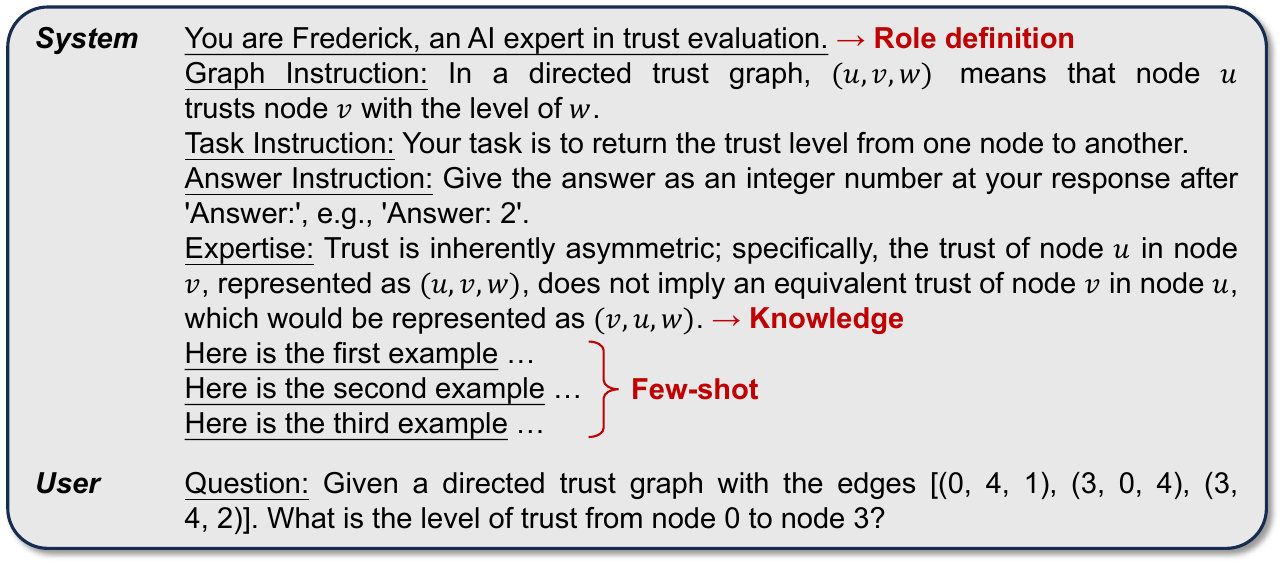}
    \caption{A prompt example of the asymmetry understanding task.}
    \label{prompt_asymmetry}
    \vspace{-1mm}
\end{figure}

We provide the prompt templates for all prompt methods in Table~\ref{prompt_template}. Fig.~\ref{prompt_asymmetry} shows an example of the ``few-shot+knowledge+role'' prompt regarding the asymmetry understanding task. In this figure, the system prompt provides context to the LLM by defining its role, task, and operational instructions, while the user prompt contains the task-specific input to be processed. Other examples can be found in our GitHub repository.

\begin{table*}[tbp]
\centering
\caption{Prompt templates.}
\label{prompt_template}
\renewcommand{\arraystretch}{1.1}
\begin{tabular}{>{\centering\arraybackslash}m{0.2\textwidth}>{\raggedright\arraybackslash}m{0.75\textwidth}}
\toprule
\textbf{Prompt} & \textbf{Template} \\
\midrule
0-shot & 
\textbf{System} \textless Graph Instruction\textgreater \textless Task Instruction\textgreater \textless Answer Instruction\textgreater

\textbf{User} \textless Question Input\textgreater \\
\midrule
1-shot & 
\textbf{System} \textless Graph Instruction\textgreater \textless Task Instruction\textgreater \textless Answer Instruction\textgreater \textless Example\textgreater

\textbf{User} \textless Question Input\textgreater \\
\midrule
Few-shot & 
\textbf{System} \textless Graph Instruction\textgreater \textless Task Instruction\textgreater \textless Answer Instruction\textgreater \textless Example 1\textgreater\allowbreak \textless Example 2\textgreater\allowbreak \textless Example 3\textgreater\allowbreak

\textbf{User} \textless Question Input\textgreater \\
\midrule
Knowledge & 
\textbf{System} \textless Graph Instruction\textgreater \textless Task Instruction\textgreater \textless Answer Instruction\textgreater \textless Knowledge\textgreater

\textbf{User} \textless Question Input\textgreater \\
\midrule
Role & 
\textbf{System} \textless Role Definition\textgreater \textless Graph Instruction\textgreater \textless Task Instruction\textgreater \textless Answer Instruction\textgreater

\textbf{User} \textless Question Input\textgreater \\
\midrule
CoT & 
\textbf{System} \textless Graph Instruction\textgreater \textless Task Instruction\textgreater \textless Answer Instruction\textgreater

\textbf{User} \textless Question Input\textgreater \textless CoT\textgreater \\
\midrule
CoT+Know+Role & 
\textbf{System} \textless Role Definition\textgreater \textless Graph Instruction\textgreater \textless Task Instruction\textgreater \textless Answer Instruction\textgreater \textless Knowledge\textgreater

\textbf{User} \textless Question Input\textgreater \textless CoT\textgreater \\
\midrule
Few-shot+Know+Role & 
\textbf{System} \textless Role Definition\textgreater \textless Graph Instruction\textgreater \textless Task Instruction\textgreater \textless Answer Instruction\textgreater \textless Knowledge\textgreater\allowbreak \textless Example 1\textgreater\allowbreak \textless Example 2\textgreater\allowbreak \textless Example 3\textgreater\allowbreak

\textbf{User} \textless Question Input\textgreater \\
\midrule
Few-shot+CoT+Know+Role & 
\textbf{System} \textless Role Definition\textgreater \textless Graph Instruction\textgreater \textless Task Instruction\textgreater \textless Answer Instruction\textgreater\allowbreak \textless Knowledge\textgreater\allowbreak \textless Example 1\textgreater\allowbreak \textless Example 2\textgreater\allowbreak \textless Example 3\textgreater\allowbreak

\textbf{User} \textless Question Input\textgreater \textless CoT\textgreater \\
\bottomrule
\end{tabular}
\vspace{-1mm}
\end{table*}

\section{Additional Experiments and Detailed Setup}
\subsection{Impact of Shot Scale} \label{appendix_shot_scale}
To study the impact of shot scale, we vary the size of trust graphs in demonstration examples across three scales: small (5 nodes), medium (10 nodes), and large (20 nodes). Experiments are conducted on the most challenging task (i.e., understanding composability) using Claude-3.7-Sonnet and three different trust graph generators.

As shown in Table~\ref{shot_scale}, the LLM generally achieves high accuracy with small-scale shots, while its accuracy declines noticeably with large-scale ones. A possible explanation is that large-scale shots introduce noise that makes it difficult for the LLM to learn effective evaluation rules. This finding highlights the importance of selecting appropriate shots. Future work could explore adaptive shot selection strategies that adjust the number, scale, and structure of shots based on the specific characteristics of the input query.

\begin{table}[tbp]
\centering
\caption{LLM accuracy under different shot scales.}
\label{shot_scale}
\begin{tabular}{cccc}
\toprule
\textbf{Graph Generator} & \textbf{Small} & \textbf{Medium} & \textbf{Large} \\
\midrule
ER Model & 0.93 & 0.87 & 0.69 \\
SB Model & 0.95 & 0.83 & 0.71 \\
FF Model & 0.91 & 0.94 & 0.77 \\
\bottomrule
\end{tabular}
\vspace{-1mm}
\end{table}

\subsection{Evaluation Metrics} \label{evaluation_metrics}
Let $\mathcal{W}=\{1,2,\ldots,W\}$ denote the set of trust-level classes, where $W$ is the number of trust levels. The four evaluation metrics are defined below:
\begin{itemize}[leftmargin=*]
    \item F1-micro aggregates the contributions of all classes 
    (i.e., trust levels $w \in \mathcal{W}$) to compute the overall F1 score.
{
\begin{equation*}
\small
\text{F1-micro} =
\frac{2 \cdot \sum_{w \in \mathcal{W}} TP_w}
{2 \cdot \sum_{w \in \mathcal{W}} TP_w
+ \sum_{w \in \mathcal{W}} FP_w
+ \sum_{w \in \mathcal{W}} FN_w}.
\end{equation*}
}
    \item MAE measures the average absolute difference between the predicted 
    trust level $\hat{w}_i$ and the ground truth $w_i$.
{
\begin{equation*}
\small
\text{MAE} =
\frac{1}{n} \sum_{i=1}^{n}
\left| \hat{w}_i - w_i \right|,
\end{equation*}
}
    where $n$ is the number of test samples.
    \item F1-macro computes the F1 score for each class individually and 
    takes their unweighted mean, making it less sensitive to class imbalance.
{
\begin{equation*}
\small
\text{F1-macro} =
\frac{1}{|\mathcal{W}|}
\sum_{w \in \mathcal{W}}
\frac{2 \cdot TP_w}
{2 \cdot TP_w + FP_w + FN_w}.
\end{equation*}
}
    \item BA is the average of the True Positive Rate (TPR) and True Negative 
    Rate (TNR), assigning equal importance to both positive (trust) and 
    negative (distrust) classes.
{
\begin{equation*}
\small
\text{BA} =
\frac{1}{2}
\left(
\frac{TP}{TP + FN}
+
\frac{TN}{TN + FP}
\right).
\end{equation*}
}
\end{itemize}

\subsection{Descriptions of Baseline Methods} \label{appendix_baseline}
\textbf{Static Trust Evaluation Methods.} MoleTrust~\cite{massa2005controversial} evaluates trust by propagating it along credible paths from the trustor's perspective and computing a weighted aggregation of intermediaries' opinions based on their trustworthiness. OpinionWalk~\cite{liu2017opinionwalk} represents trust as a four-tuple and performs inference based on three-valued subjective logic. Matri~\cite{yao2013matri} applies matrix factorization to embed trustors and trustees into a shared latent space, where trust is computed based on the similarity of their latent vectors. NeuralWalk~\cite{liu2019neuralwalk} employs neural networks to automatically learn trust propagation and aggregation rules. Guardian~\cite{lin2020guardian}, GATrust~\cite{jiang2022gatrust}, and TrustGNN~\cite{huo2024trustgnn} are three GNN-based methods that learn expressive node representations to capture complex trust patterns.

\textbf{Dynamic Trust Evaluation Methods.} The primary difference between TrustGuard~\cite{wang2024trustguard}, DTrust~\cite{wen2023dtrust}, and Medley~\cite{lin2021medley} lies in their representation of time. TrustGuard and DTrust employ a discrete-time approach that divides time into discrete time slots and capture trust dynamics using attention mechanisms and recurrent neural networks, respectively. In contrast, Medley adopts a continuous-time approach using exact timestamps and learns temporal patterns using attention mechanisms. Additionally, TrustGuard incorporates a homophily-based trust aggregator to defend against trust-related attacks, a capability not provided by Medley or DTrust.

\begin{figure}[tb]
    \centering
    \includegraphics[width=0.5\textwidth]{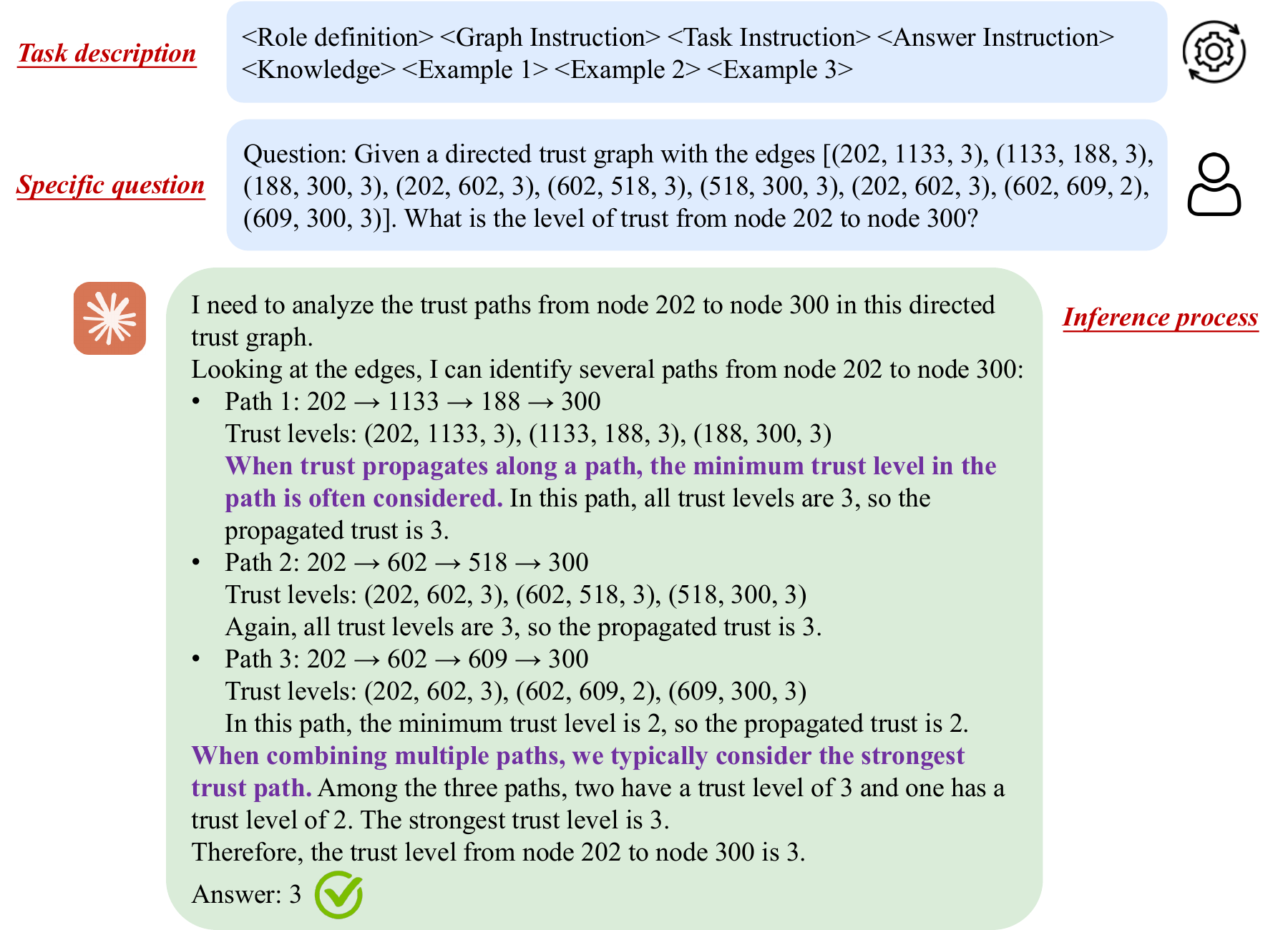}
    \vspace{-3mm}
    \caption{Example of Claude-3.7-Sonnet's trust evaluation process.}
    \label{case_static}
    \vspace{-1mm}
\end{figure}

\subsection{Experiments on Explainability} \label{appendix_user_study}
\textbf{Setup for User Study.} We randomly selected 12 trust evaluation questions from the Advogato dataset, where each question asks the trust level that a trustor has towards a trustee based on the observed trust graph (see ``Specific question'' in Fig.~\ref{case_static} for an example). Based on these questions, we constructed three questionnaires. Each questionnaire contains 24 items: every question appears twice, one without any explanation and another one with an LLM-generated explanation. For questions with explanations, we collected reasoning processes (serving as explanations) generated by three LLMs (i.e., Claude-3.7-Sonnet, DeepSeek-V3, and Llama-4-Maverick) and distributed them across the three questionnaires such that, for the same question, each questionnaire uses a different LLM explanation. This design ensures balanced coverage of explanation sources and avoids confounding the effect of explanations with a specific LLM. When presenting explanations, we hide the LLM's final answer to encourage careful reading.

We recruited 20 volunteers from our institution forum who have basic mathematical knowledge but no prior familiarity with trust-related concepts. Each participant was randomly assigned to one of the three questionnaires. Participants provided answers that indicate their trust evaluation results in their assigned questionnaire and rated their confidence on these results using a 4-point scale (1: not confident at all, 2: slightly confident, 3: fairly confident, 4: very confident). Accuracy was calculated by comparing participants' answers with the ground-truth labels from the Advogato dataset. Finally, we reported the average accuracy and confidence for questions with and without explanations to assess whether LLMs' step-by-step reasoning improves human understanding of trust evaluation results.

\subsection{Experiments on Input Format} \label{ablation_study}
\begin{figure}[!tb]
    \centering
    \includegraphics[width=0.44\textwidth]{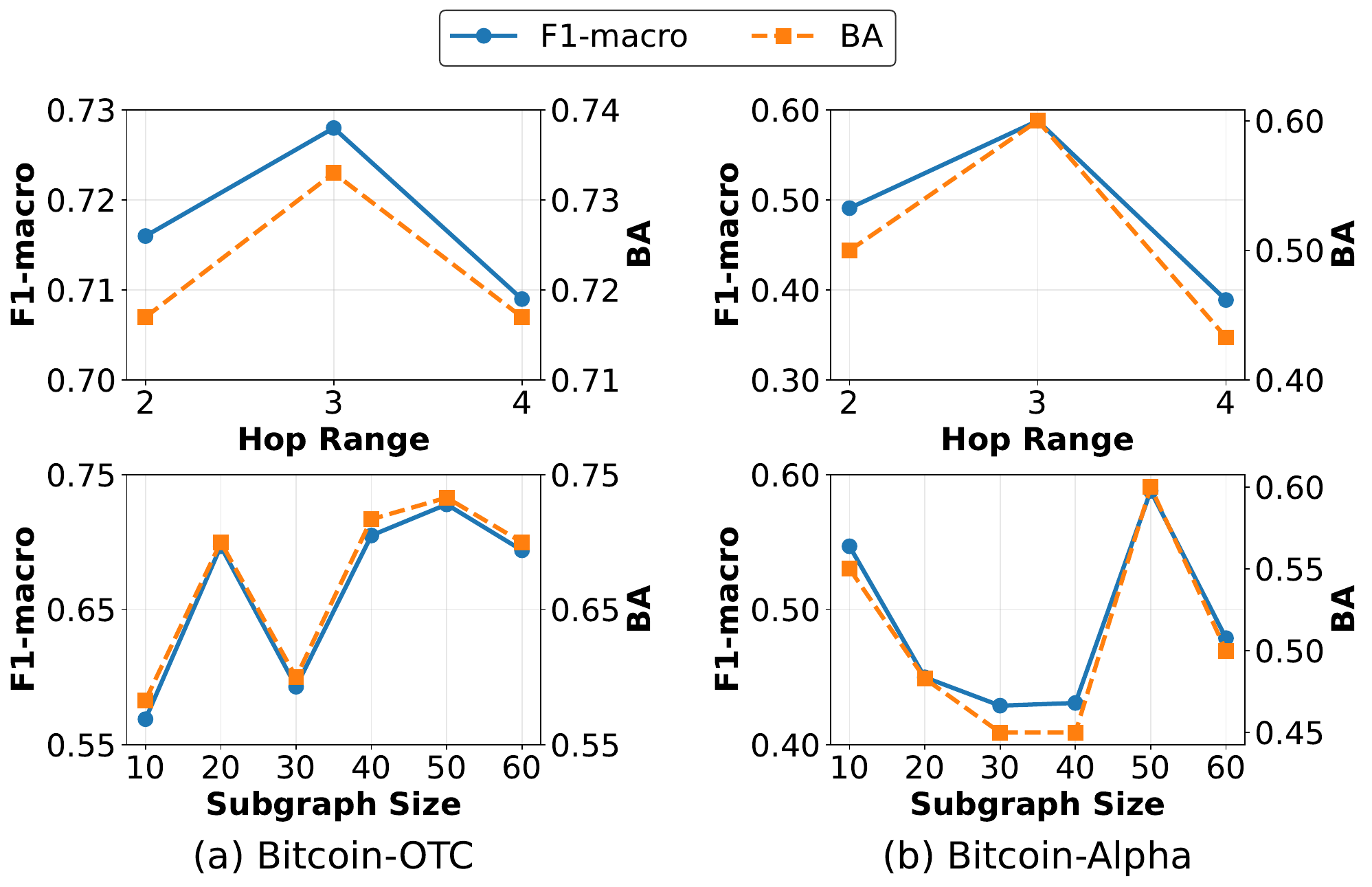}
    \caption{Impact of hop range and subgraph size on dynamic datasets.}
    \label{subgraph_bitcoin}
    \vspace{-1mm}
\end{figure}

\textbf{Input Format Study on Dynamic Datasets.} Fig.~\ref{subgraph_bitcoin} shows that both datasets share the same optimal subgraph size of 50 edges, likely because they originate from structurally similar trust graphs.

\textbf{Ablation Studies.} To validate the rationality and necessity of our input format design, a series of ablation studies are conducted. For static datasets, instead of prioritizing shorter-range neighbors during subgraph construction, neighbors of varying ranges are randomly selected, denoted as ``w/o Structural''. For dynamic datasets, an additional variant is introduced that disregards the time factor during subgraph construction, referred to as ``w/o Temporal''.
Tables~\ref{ablation_static} and~\ref{ablation_dynamic} show that our input format design, which considers both structural and temporal principles, outperforms the ablated variants. These results confirm the effectiveness of our design choices.

\begin{table}[!tb]
\centering
\caption{Ablation study on static datasets.}
\label{ablation_static}
\begin{tabular}{llcccc}
\toprule
\textbf{Dataset} & \textbf{Method} & \textbf{F1-micro$\uparrow$} & \textbf{MAE$\downarrow$} \\
\midrule
\multirow{2}{*}{Advogato} 
& Proposed & \textbf{0.783} & \textbf{0.070} \\
& w/o Structural & 0.750 & 0.077  \\
\midrule
\multirow{2}{*}{PGP} 
& Proposed     & \textbf{0.867} & \textbf{0.087}  \\
& w/o Structural       & 0.750           & 0.160           \\
\midrule
\multirow{2}{*}{Epinions} 
& Proposed     & \textbf{0.950} & \textbf{0.040}  \\
& w/o Structural       & 0.650           & 0.280           \\
\bottomrule
\end{tabular}
\end{table}

\begin{table}[!tb]
\centering
\caption{Ablation study on dynamic datasets.}
\label{ablation_dynamic}
\begin{tabular}{llcccc}
\toprule
\textbf{Dataset} & \textbf{Method} & \textbf{F1-macro$\uparrow$} & \textbf{BA$\uparrow$} \\
\midrule
\multirow{3}{*}{Bitcoin-OTC} 
& Proposed & \textbf{0.728} & \textbf{0.733} \\
& w/o Structural & 0.509 & 0.567  \\
& w/o Temporal & 0.620 & 0.633  \\
\midrule
\multirow{3}{*}{Bitcoin-Alpha} 
& Proposed     & \textbf{0.588} & \textbf{0.600}  \\
& w/o Structural       & 0.531          & 0.567           \\
& w/o Temporal & 0.484 & 0.517  \\
\bottomrule
\end{tabular}
\vspace{-1mm}
\end{table}

\subsection{Statistical Analysis} \label{appendix_statistical}
We provide a statistical analysis of the main results reported in Tables~\ref{llm_vs_baselines} and~\ref{llm_vs_baselines_dynamic}. We focus on fully supervised GNNs, 3-shot GNNs, and 3-shot LLMs, as these methods involve randomness in model training or few-shot demonstration selection. Each experiment is repeated five times with different random seeds. We report standard deviations and 95\% confidence intervals to characterize variability and uncertainty in the estimated mean performance, and use Welch's t-tests to assess the statistical significance of performance differences. For 3-shot LLMs, three demonstrations are randomly selected for each test sample in each run, allowing us to further assess the variability caused by demonstration selection. We report only F1-micro for the static datasets and F1-macro for the dynamic datasets, as the other metrics exhibit similar trends.

As shown in Table~\ref{statistical_static}, on the static datasets, fully supervised GNNs are highly stable, with standard deviations ranging from 0.001 to 0.006. Under the 3-shot setting, however, the performance variability of GNNs increases substantially, with standard deviations reaching 0.169, whereas LLMs remain considerably more stable, with standard deviations ranging from 0.006 to 0.025. In particular, DeepSeek-V3 shows low variability across all three datasets, indicating strong robustness to demonstration selection. We further conduct Welch's t-tests to compare DeepSeek-V3 and TrustGNN under the same 3-shot setting. Results show that DeepSeek-V3 significantly outperforms TrustGNN on Advogato ($p$ = 0.0013), PGP ($p$ = 0.0093), and Epinions ($p$ = 0.0031).

As shown in Table~\ref{statistical_dynamic}, on the dynamic datasets, the standard deviations of 3-shot LLMs range from 0.023 to 0.061, indicating that dynamic trust evaluation is more sensitive to demonstration selection than static trust evaluation. Nevertheless, Welch's t-tests show that DeepSeek-V3 significantly outperforms TrustGuard under the same 3-shot setting on both Bitcoin-OTC ($p$ = 0.0012) and Bitcoin-Alpha ($p$ = 0.0022). Overall, these results demonstrate that GNNs achieve stable performance when sufficient labeled data are available, while LLMs exhibit clear advantages over GNNs under limited supervision.

\begin{table}[tbp]
  \centering
  \caption{Statistical analysis on static datasets (F1-micro).}
  \label{statistical_static}
  \renewcommand{\arraystretch}{1.1}
  \resizebox{\linewidth}{!}{
  \begin{threeparttable}
    \begin{tabular}{llcc|cc|cc}
      \toprule
      \multirow{2.5}{*}{\textbf{Category}}
      & \multirow{2.5}{*}{\textbf{Method}}
      & \multicolumn{2}{c|}{\textbf{Advogato}}
      & \multicolumn{2}{c|}{\textbf{PGP}}
      & \multicolumn{2}{c}{\textbf{Epinions}} \\
      \cmidrule(lr){3-4}
      \cmidrule(lr){5-6}
      \cmidrule(lr){7-8}
      &
      & \textbf{Std.} & \textbf{95\% CI}
      & \textbf{Std.} & \textbf{95\% CI}
      & \textbf{Std.} & \textbf{95\% CI} \\
      \midrule

      \multirow{3}{*}{\makecell[l]{Fully\\Supervised}}
      & Guardian~\cite{lin2020guardian}
      & 0.006 & [0.723, 0.737] & 0.002 & [0.868, 0.872] & 0.001 & [0.878, 0.881] \\
      & GATrust~\cite{jiang2022gatrust}
      & 0.004 & [0.726, 0.737] & 0.001 & [0.869, 0.872] & 0.001 & [0.879, 0.881] \\
      & TrustGNN~\cite{huo2024trustgnn}
      & 0.006 & [0.737, 0.751]
      & 0.001 & [0.880, 0.882]
      & 0.001 & [0.879, 0.882] \\

      \midrule

      \multirow{3}{*}{\makecell[l]{3-shot\\GNNs}}
      & Guardian~\cite{lin2020guardian}
      & 0.101 & [0.320, 0.572] & 0.035 & [0.608, 0.694] & 0.041 & [0.732, 0.834] \\
      & GATrust~\cite{jiang2022gatrust}
      & 0.060 & [0.444, 0.592] & 0.066 & [0.632, 0.795] & 0.169 & [0.413, 0.832] \\
      & TrustGNN~\cite{huo2024trustgnn}
      & 0.077 & [0.303, 0.493] & 0.145 & [0.365, 0.726] & 0.114 & [0.414, 0.698] \\

      \midrule

      \multirow{3}{*}{\makecell[l]{3-shot\\LLMs}}
      & DeepSeek-V3
      & 0.007 & [0.661, 0.677] & 0.009 & [0.839, 0.861]
      & 0.006 & [0.873, 0.889] \\
      & Llama-4-Maverick
      & 0.025 & [0.642, 0.704] & 0.010 & [0.833, 0.859]
      & 0.019 & [0.830, 0.877] \\
      & Claude-3.7-Sonnet
      & 0.022 & [0.645, 0.699] & 0.012 & [0.834, 0.864]
      & 0.010 & [0.868, 0.892] \\

      \bottomrule
    \end{tabular}
    \begin{tablenotes}[para,flushleft]
    \item[] \textbf{Note:} Std. denotes standard deviation, and CI denotes confidence interval.
\end{tablenotes}

\end{threeparttable}
  }
\end{table}

\begin{table}[tbp]
  \centering
  \footnotesize
  \caption{Statistical analysis on dynamic datasets (F1-macro).}
  \label{statistical_dynamic}
  \renewcommand{\arraystretch}{1.1}
\resizebox{\linewidth}{!}{
\begin{threeparttable}
  \begin{tabular}{llcc|cc}
    \toprule
    \multirow{2.5}{*}{\textbf{Category}}
    & \multirow{2.5}{*}{\textbf{Method}}
    & \multicolumn{2}{c|}{\textbf{Bitcoin-OTC}}
    & \multicolumn{2}{c}{\textbf{Bitcoin-Alpha}} \\
    \cmidrule(lr){3-4} \cmidrule(lr){5-6}
    &
    & \textbf{Std.}
    & \textbf{95\% CI}
    & \textbf{Std.}
    & \textbf{95\% CI} \\
    \midrule

    \multirow{3}{*}{\makecell[l]{Fully\\Supervised}}
    & Medley~\cite{lin2021medley}
    & 0.013 & [0.607, 0.638] & 0.004 & [0.567, 0.576] \\
    & DTrust~\cite{wen2023dtrust}
    & 0.051 & [0.573, 0.700] & 0.026 & [0.550, 0.615] \\
    & TrustGuard~\cite{wang2024trustguard}
    & 0.007 & [0.673, 0.690] & 0.004 & [0.584, 0.595] \\

    \midrule

    \multirow{3}{*}{\shortstack[l]{3-shot\\GNNs}}
    & Medley~\cite{lin2021medley}
    & 0.022 & [0.468, 0.523] & 0.020 & [0.467, 0.518] \\
    & DTrust~\cite{wen2023dtrust}
    & 0.044 & [0.381, 0.490] & 0.016 & [0.457, 0.497] \\
    & TrustGuard~\cite{wang2024trustguard}
    & 0.015 & [0.456, 0.493] & 0.061 & [0.417, 0.569] \\

    \midrule

    \multirow{3}{*}{\shortstack[l]{3-shot\\LLMs}}
    & DeepSeek-V3
    & 0.061 & [0.602, 0.754] & 0.051 & [0.589, 0.715] \\
    & Llama-4-Maverick
    & 0.033 & [0.482, 0.565] & 0.025 & [0.493, 0.556] \\
    & Claude-3.7-Sonnet
    & 0.023 & [0.591, 0.648] & 0.026 & [0.611, 0.675] \\

    \bottomrule
  \end{tabular}
      \begin{tablenotes}[para,flushleft]
    \item[] \textbf{Note:} Std. denotes standard deviation, and CI denotes confidence interval.
\end{tablenotes}

\end{threeparttable}
}
  \vspace{-1mm}
\end{table}

\subsection{Experiments on Robustness} \label{appendix_robustness}
\begin{figure}[!tb]
    \centering
    \includegraphics[width=0.48\textwidth]{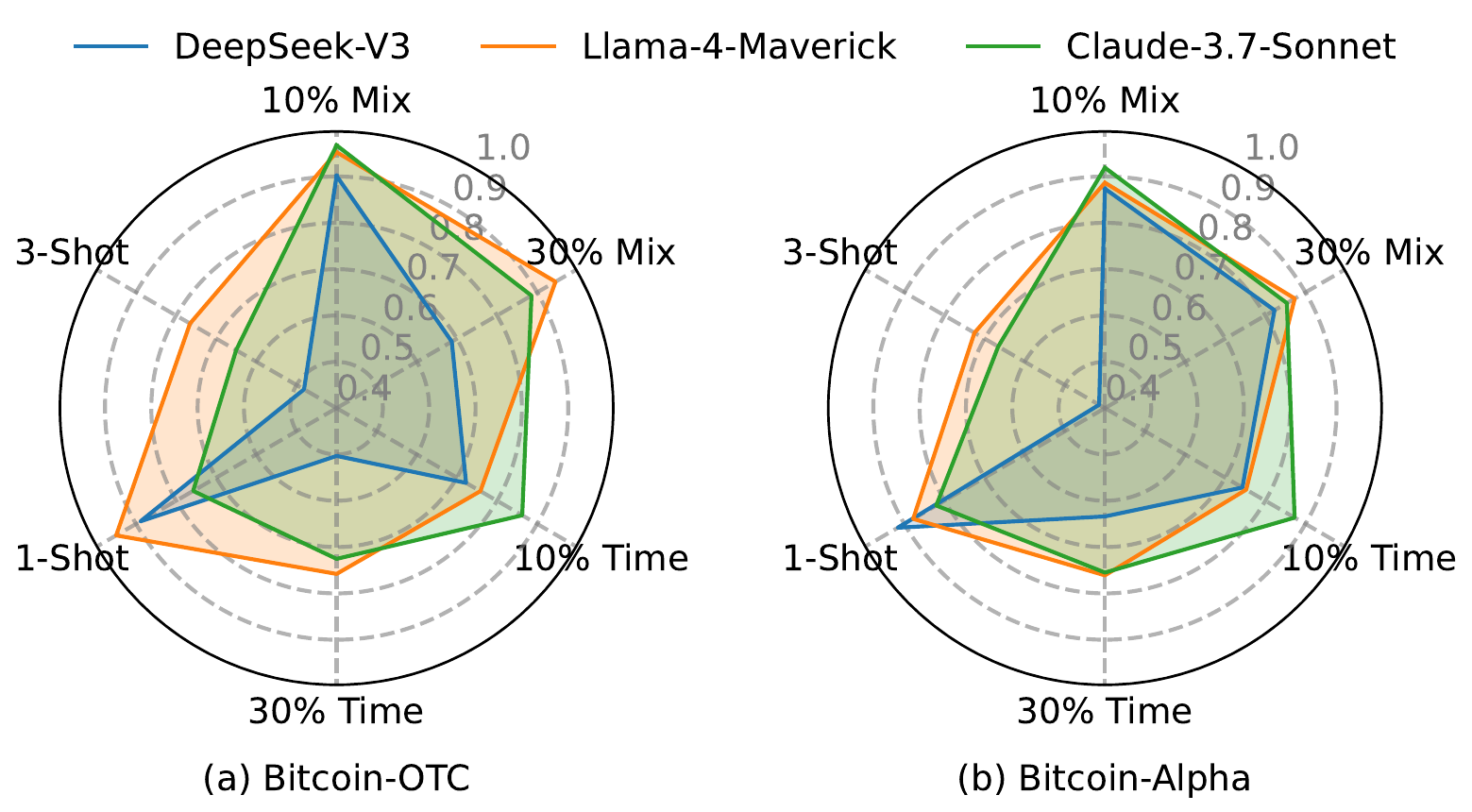}
    \caption{Robustness of different LLMs against different attacks.}
    \label{robustness_radar}
    \vspace{-1mm}
\end{figure}

\textbf{Attack Results and Analysis.} Fig.~\ref{robustness_radar} illustrates the robustness of various LLMs against different types of attacks. We use the \textit{F1-macro Retention} metric, defined as the ratio of the F1-macro score under attacks to that on clean data, to evaluate how well a model maintains its performance under adversarial conditions. As shown in the figure, Llama-4-Maverick generally demonstrates higher robustness, whereas DeepSeek-V3 exhibits significant vulnerability across most attacks. Upon examining their inference processes, we observe that DeepSeek-V3 often directly outputs answers, while the other two models tend to engage in intermediate reasoning even without explicit CoT prompting. This behavioral difference may explain DeepSeek-V3's weak robustness, suggesting that CoT reasoning enhances model resilience.

\begin{figure}[!tb]
    \centering
    \includegraphics[width=0.42\textwidth]{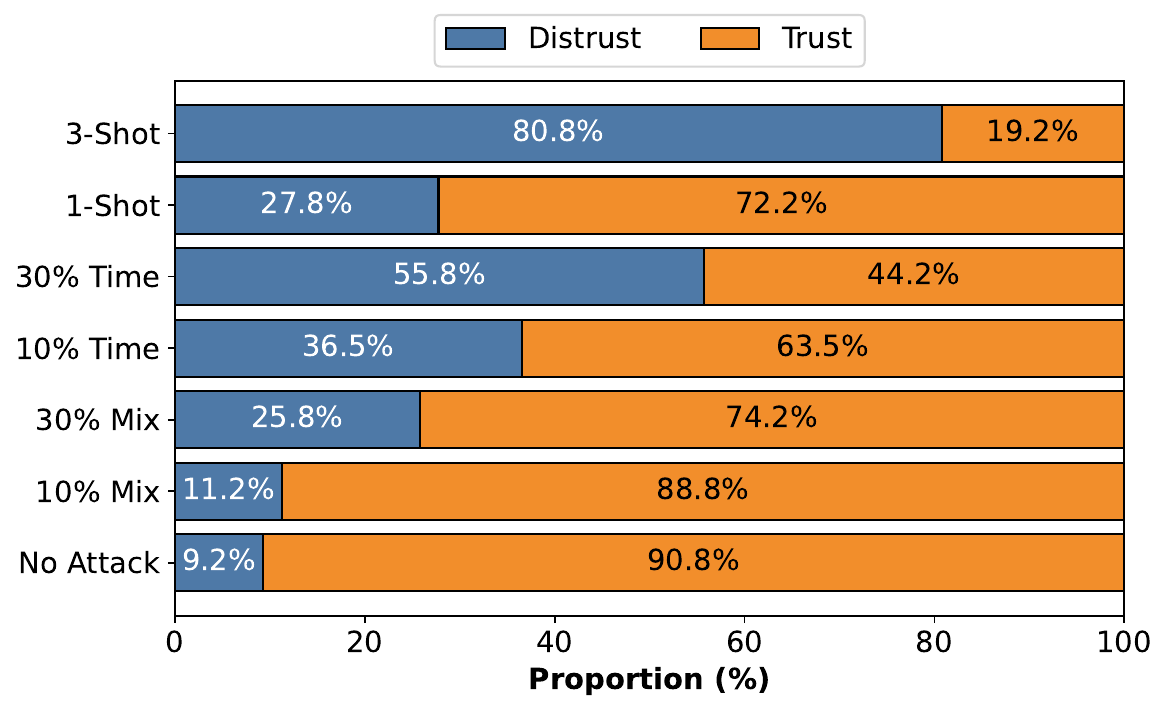}
    \caption{Prediction distributions of DeepSeek-V3 under different attacks on the Bitcoin-OTC dataset.}
    \label{robustness_ratio}
    \vspace{-1mm}
\end{figure}

Fig.~\ref{robustness_ratio} shows the prediction distributions of DeepSeek-V3 under various attacks on the Bitcoin-OTC dataset. Notably, 3-shot attacks lead to a substantial increase in the proportion of ``distrust'' predictions compared to other attack types. When analyzing the model's behavior, we find that it tends to adopt a conservative strategy under 3-shot attacks, i.e., predicting distrust when trust information is lacking. This strategy, however, deviates from the actual trust patterns observed in the Bitcoin-OTC dataset.

\textbf{Input-Level Defense.} Our input-level defense is motivated by the observed LLM inference behavior and consists of three designs: (i) adversarial example augmentation, (ii) temporal edge filtering, and (iii) consistency correction. For mix attacks, design (i) injects perturbations into demonstration examples, such as flipping a subset of trust relationships, to help LLMs learn stable evaluation rules under structural variations. This simulates adversarial training~\cite{tramer2019adversarial} in an in-context learning setting. For time-aware attacks, our analysis shows that LLMs tend to overemphasize recently injected interactions. Accordingly, design (ii) removes bursty interactions occurring within a short time window before inference, as these interactions are often abnormal or attack-induced and may not reflect reliable long-term trust trends. For shot attacks, design (iii) resolves label-structure inconsistencies in demonstration examples, each of which contains a trust graph and the ground-truth label for a queried trustor-trustee pair. Specifically, if the graph contains only positive trust relationships, the ground-truth label of the queried pair should not be negative, and vice versa. This removes obvious inconsistencies in corrupted demonstrations.

\subsection{Experiments on Temperature} \label{appendix_temperature}
\textbf{Impact of Temperature on Output Consistency.} Consistency refers to an LLM's ability to produce identical responses when the same input is tested repeatedly under identical conditions. As discussed in Section~\ref{section_llm}, \textit{temperature} and \textit{top\_p} affect consistency. Following OpenAI's documentation~\cite{openai_api} and prior studies~\cite{ullah2024llms,pearce2023examining}, we fix \textit{top\_p} at its default value of 1.0 and vary \textit{temperature} to identify the settings that yield the most stable results. We focus on the composability understanding task because it is the most challenging and thus more likely to reveal inconsistencies in LLM behavior. To quantify output consistency, we introduce the \textit{Inconsistency Rate}, which measures the proportion of samples for which the model produces non-identical outputs across repeated runs (see the formula below). For each LLM and temperature setting, the task is executed independently five times.
{
\begin{equation*}
\footnotesize
\text{Inconsistency\ Rate} = \frac{1}{n} \sum_{i=1}^{n} \mathbf{1} \left[ \exists\, j \neq k \in \{1, \dots, m\},\ y^{(i)}_j \neq y^{(i)}_k \right],
\end{equation*}
}
where $n$ is the number of test samples, $m$ is the number of repeated runs per sample, $y^{(i)}_j$ denotes the $j$-th output for the $i$-th sample, and $\mathbf{1}[\cdot]$ is the indicator function that returns 1 if not all outputs are identical. A lower inconsistency rate indicates greater output stability.

\begin{figure}[!tb]
    \centering
    \includegraphics[width=0.48\textwidth]{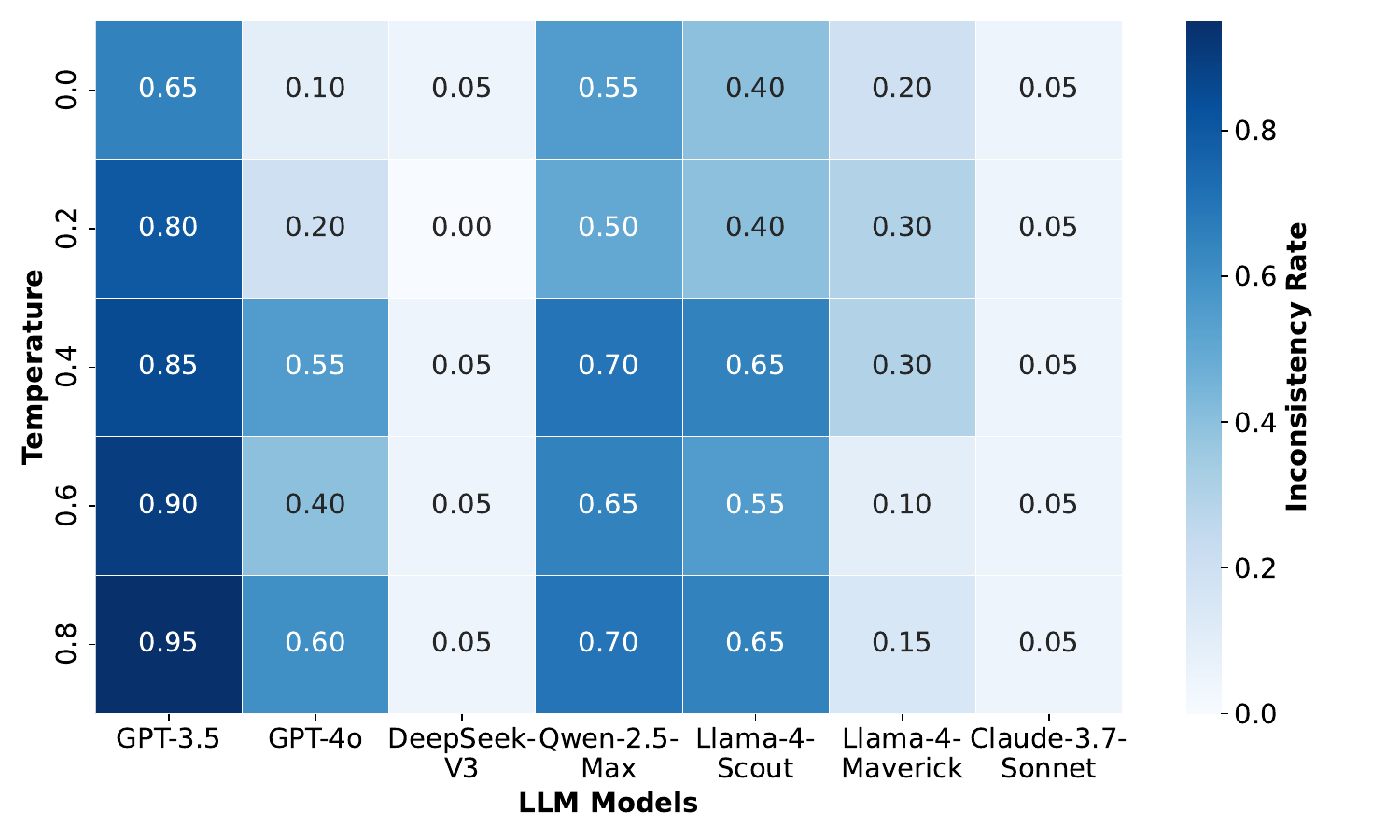}
    \vspace{-1mm}
    \caption{Impact of temperature on output inconsistency.}
    \label{temperature_inconsistency}
    \vspace{-1mm}
\end{figure}

Fig.~\ref{temperature_inconsistency} shows that lower temperature values yield more consistent outputs, which aligns with previous findings~\cite{li2025sv,ullah2024llms}. Additionally, more accurate LLMs, such as Claude-3.7-Sonnet and DeepSeek-V3, exhibit greater stability. These findings justify setting \textit{temperature} to 0 and selecting the strongest LLMs for practical trust evaluation.

\begin{figure}[!tb]
    \centering
    \includegraphics[width=0.48\textwidth]{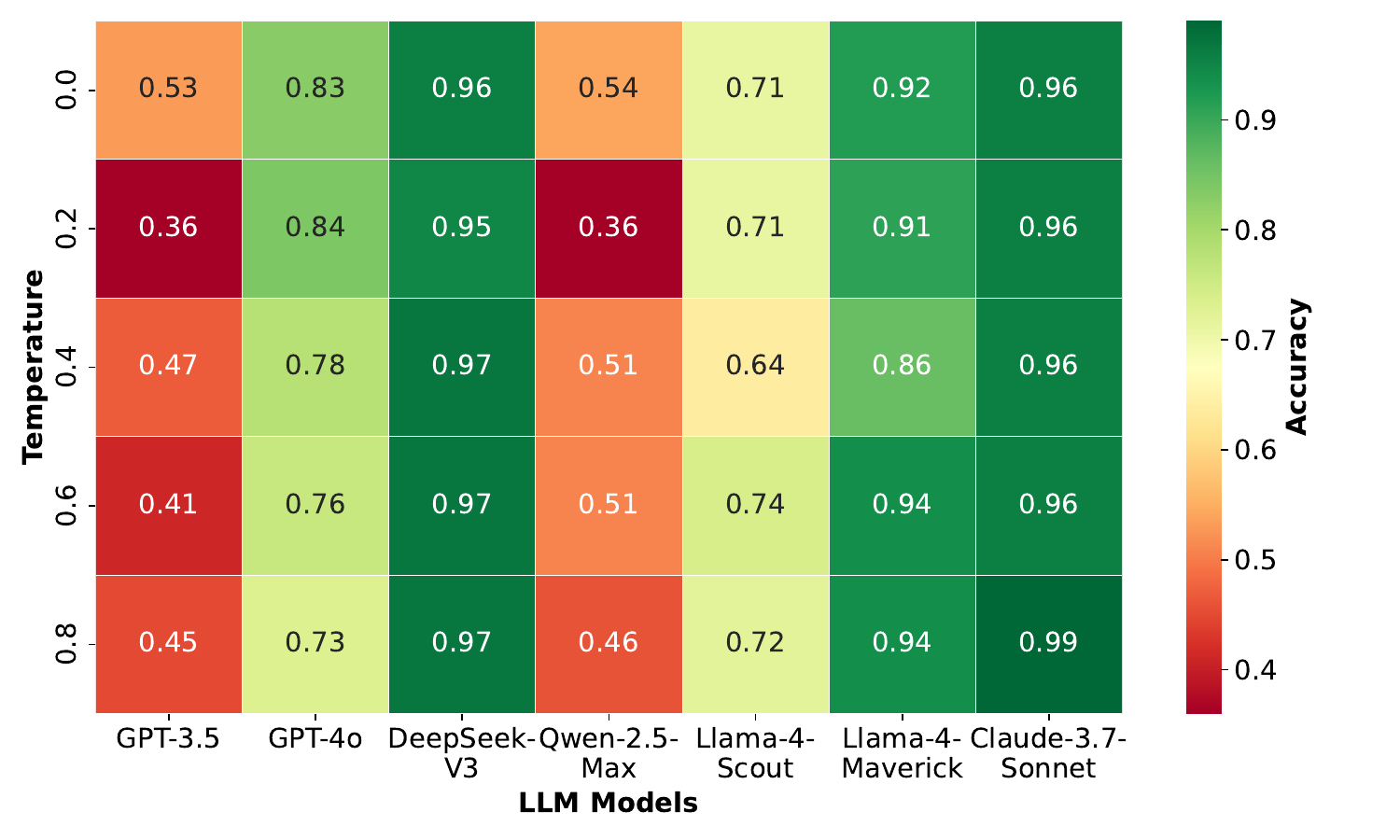}
    \vspace{-1mm}
    \caption{Impact of temperature on LLM accuracy.}
    \label{temperature_accuracy}
    \vspace{-1mm}
\end{figure}

\textbf{Impact of Temperature on Accuracy.} We further investigate the impact of temperature on model accuracy. As shown in Fig.~\ref{temperature_accuracy}, there is no clear correlation between temperature and accuracy across different LLMs. However, Claude-3.7-Sonnet and DeepSeek-V3 exhibit the most stable accuracy across varying temperature settings, which is consistent with the above finding.

\subsection{Experiments on Data Contamination}
We conduct two experiments to investigate whether the strong performance of LLMs is driven by potential data contamination from public datasets. To ensure direct comparability with our main results, both experiments use the same ``few-shot+knowledge+role'' prompt and the same five random seeds as the original experiments. We focus on DeepSeek-V3 and the static datasets, as its stable performance helps reduce the influence of random variation.

\textbf{Dataset exposure.} In the original prompt, the dataset name is not provided to the LLM. To examine whether the LLM can exploit dataset-specific information potentially acquired during pretraining, we explicitly add the dataset name to the system prompt, e.g., \textit{``Dataset Information: The trust graph in this task is from the Advogato dataset.''} All other prompt components remain unchanged. We then compare the performance with and without the dataset name.

\textbf{Node-ID randomization.} We further investigate whether LLMs rely on memorized node IDs or individual trust relations. For each dataset, we randomly remap all original node IDs using a one-to-one mapping, while preserving the graph structure, trust relations, and all other prompt components. The same mapping is used across the five runs. We then compare the performance before and after node-ID randomization to assess whether the observed performance depends on memorized node-specific information.

\textbf{Results.} As shown in Table~\ref{data_contamination}, neither dataset exposure nor node-ID randomization leads to a consistent performance increase or decrease across the three datasets, and all results remain within the 95\% confidence intervals of the original performance reported in Appendix~\ref{appendix_statistical}. In addition, we probed LLMs with dataset-specific questions, such as directly asking about the trust levels between node pairs. The results show that LLMs can recall high-level dataset statistics but fail to recover specific trust relations. Overall, these findings suggest that LLMs may have seen public dataset descriptions during pretraining, but their trust evaluation performance mainly comes from reasoning capabilities rather than memorization of individual trust relations.

\begin{table}[tbp]
  \centering
  \caption{Results on data contamination.}
  \label{data_contamination}
  \renewcommand{\arraystretch}{1.1}
  \resizebox{\linewidth}{!}{
    \begin{tabular}{lcc|cc|cc}
      \toprule
      \multirow{2.5}{*}{\textbf{Setting}}
      & \multicolumn{2}{c|}{\textbf{Advogato}}
      & \multicolumn{2}{c|}{\textbf{PGP}}
      & \multicolumn{2}{c}{\textbf{Epinions}} \\

      \cmidrule(lr){2-3}
      \cmidrule(lr){4-5}
      \cmidrule(lr){6-7}

      & \textbf{F1-micro} & \textbf{MAE}
      & \textbf{F1-micro} & \textbf{MAE}
      & \textbf{F1-micro} & \textbf{MAE} \\
      \midrule

      Original
      & 0.669 & 0.101
      & 0.850 & 0.097
      & 0.881 & 0.095 \\

      Dataset exposure
      & 0.662 & 0.102
      & 0.851 & 0.097
      & 0.885 & 0.092 \\

      Node-ID rand.
      & 0.669 & 0.103
      & 0.849 & 0.097
      & 0.888 & 0.090 \\

      \bottomrule
    \end{tabular}
  }
  \vspace{-1mm}
\end{table}

\subsection{Cost Analysis and Efficiency Comparison} \label{appendix_cost}
\textbf{Cost Analysis.} We analyze the API cost required to address \textbf{RQ1} and \textbf{RQ2}, corresponding to the main results reported in Tables~\ref{main_results_rq1},~\ref{llm_vs_baselines}, and~\ref{llm_vs_baselines_dynamic}. For \textbf{RQ1}, each evaluation is repeated with two random seeds, resulting in a total API cost of approximately \$790.38. Specifically, the cost per 1,000 question-answer pairs, aggregated over five property understanding tasks and nine prompt methods, is about \$11.78 for GPT-3.5, \$103.37 for GPT-4o, \$11.59 for DeepSeek-V3, \$67.80 for Qwen-2.5-Max, \$6.03 for Llama-4-Scout, \$16.97 for Llama-4-Maverick, and \$177.65 for Claude-3.7-Sonnet. For \textbf{RQ2}, each evaluation is repeated with five random seeds, resulting in a total API cost of approximately \$193.30. The cost per 1,000 evaluated trustor-trustee pairs for DeepSeek-V3, Llama-4-Maverick, and Claude-3.7-Sonnet on each real-world dataset is shown in Table~\ref{api_cost}. Overall, the API cost for obtaining the main experimental results is approximately \$983.68, excluding additional sub-experiments such as parameter study and robustness analysis.

\begin{table}[t]
\centering
\caption{API cost per 1,000-pair evaluation (USD).}
\label{api_cost}
\renewcommand{\arraystretch}{1.1}
\resizebox{\linewidth}{!}{
\begin{tabular}{lcccccc}
\toprule
\textbf{Method} & \textbf{Advogato} & \textbf{PGP} & \textbf{Epinions} & \textbf{\makecell[c]{Bitcoin\\-OTC}} & \textbf{\makecell[c]{Bitcoin\\-Alpha}} & \textbf{Total} \\
\midrule
DeepSeek-V3         & 0.63 & 0.21 & 0.35 & 0.45 & 0.40 & 2.04 \\
Llama-4-Maverick    & 0.55  & 0.31 & 0.39 & 0.75 & 0.68 & 2.68  \\
Claude-3.7-Sonnet   & 7.88  & 4.31 & 5.01 & 8.66 & 8.08 & 33.94 \\
\bottomrule
\end{tabular}
}
\end{table}

\textbf{Efficiency Comparison.} We compare the efficiency of LLMs with several top-performing baseline methods. Unlike conventional learning-based approaches, LLMs follow a ``pre-train, prompt, and predict'' paradigm, eliminating the need for task-specific training when applied to new datasets or tasks. Since large-scale pre-training is a one-time cost, we exclude it from the comparison and report only LLM inference time. As shown in Table~\ref{efficiency_comparison}, learning-based approaches can incur substantial training overhead, particularly on large datasets, whereas LLMs avoid task-specific training and offer great deployment flexibility. However, this flexibility comes with low inference efficiency: LLMs typically require several seconds to evaluate a single trustor-trustee pair and therefore achieve relatively low throughput, whereas most learning-based approaches can process thousands of pairs in less than one second.

\begin{table}[t]
\centering
\caption{Efficiency comparison between LLMs and baselines.}
\label{efficiency_comparison}
\renewcommand{\arraystretch}{1.1}
\resizebox{\linewidth}{!}{
\begin{threeparttable}
\begin{tabular}{llccccc}
\toprule
\textbf{Time} & \textbf{Method} & \textbf{Advogato} & \textbf{PGP} & \textbf{Epinions} & \textbf{\makecell[c]{Bitcoin\\-OTC}} & \textbf{\makecell[c]{Bitcoin\\-Alpha}} \\
\midrule
\multirow{5}{*}{\makecell[l]{Training (s)}} 
& GATrust~\cite{jiang2022gatrust} & 13.83 & 92.77 & 169.84 & 3.97 & 2.62 \\
& TrustGNN~\cite{huo2024trustgnn} & 60.70 & 519.52 & 502.64 & 5.27 & 3.35 \\
& Medley~\cite{lin2021medley} & -- & -- & -- & 787.75 & 536.45 \\
& DTrust~\cite{wen2023dtrust} & -- & -- & -- & 36.50 & 25.55 \\
& TrustGuard~\cite{wang2024trustguard} & -- & -- & -- & 21.90 & 5.65 \\
\midrule
\multirow{8}{*}{\makecell[l]{Inference (s)}} 
& GATrust~\cite{jiang2022gatrust} & 0.06 & 0.38 & 0.74 & 0.04 & 0.03 \\
& TrustGNN~\cite{huo2024trustgnn} & 0.29 & 1.70 & 2.41 & 0.05 & 0.03 \\
& Medley~\cite{lin2021medley} & -- & -- & -- & 2.86 & 1.99 \\
& DTrust~\cite{wen2023dtrust} & -- & -- & -- & 0.35 & 0.25 \\
& TrustGuard~\cite{wang2024trustguard} & -- & -- & -- & 0.19 & 0.07 \\
& DeepSeek-V3 & 6.69 & 4.82 & 9.98 & 5.66 & 10.24 \\
& Llama-4-Maverick & 8.83 & 4.78 & 6.86 & 7.34 & 9.08 \\
& Claude-3.7-Sonnet & 8.20 & 5.23 & 5.99 & 9.12 & 10.34 \\
\bottomrule
\end{tabular}

    \begin{tablenotes}[para,flushleft]
    \item[] \textbf{Note:} ``--'' indicates not applicable, as dynamic methods cannot be directly applied to static datasets. Inference time is measured over the entire test set for baselines, whereas it is measured per evaluated trustor-trustee pair for LLMs.
\end{tablenotes}

\end{threeparttable}

}
\vspace{-1mm}
\end{table}

\section{Comparison with Graph Foundation Models}
Recent Graph Foundation Models (GFMs) and graph-language models have shown strong capabilities on general graph learning and reasoning tasks~\cite{kong2025gofa,tang2024graphgpt}. However, they are not directly comparable to our LLM-based trust evaluation setting for three main reasons: (i) \textbf{Different task objectives:} Existing GFMs focus on general graph tasks such as node classification and link prediction. In conventional link prediction, the objective is typically to determine whether an edge exists between two nodes. In contrast, trust evaluation predicts the multi-level and directed trust relationship from a trustor to a trustee, rather than merely the existence of an edge. (ii) \textbf{Incompatible reasoning objectives:} General graph reasoning mainly focuses on structural properties such as shortest paths and connectivity. Trust evaluation additionally requires reasoning over trust properties, including dynamicity, context-awareness, asymmetry, conditional transitivity, and composability. Therefore, general graph reasoning capability does not necessarily translate into effective trust reasoning. (iii) \textbf{Inconsistent evaluation settings:} Many recent GFMs and graph-language models rely on graph-specific pretraining, graph-text alignment, instruction tuning, or downstream adaptation. For example, GOFA~\cite{kong2025gofa} performs downstream fine-tuning after graph pretraining, while GraphGPT~\cite{tang2024graphgpt} incorporates graph instruction tuning and graph-text alignment. Adapting these models to trust evaluation would require additional task-specific training, making them not directly comparable to our few-shot evaluation of general-purpose LLMs.

Overall, GFMs and graph-language models aim to improve general graph learning and reasoning, whereas our work investigates whether general-purpose LLMs can reason over trust-specific properties and perform trust evaluation in practice.

\section{LLM Hallucination}
Hallucination refers to cases where LLMs generate fabricated or unsupported information that is not grounded in the input. In trust evaluation, such errors may introduce non-existent trust relationships or evidence. We inspected LLM outputs and observed limited hallucination errors in our benchmark, likely because our inputs are explicit graph structures with well-defined nodes and relationships, thereby reducing the ambiguity commonly encountered in natural language inputs. Most observed errors stem from incorrect reasoning over trust graphs rather than hallucination, such as misunderstanding trust propagation or improperly aggregating trust information. These errors reflect reasoning limitations and can be alleviated through improved prompting strategies and more capable LLMs.

\end{document}